\documentclass[sigconf]{acmart}
\AtBeginDocument{%
  }

\copyrightyear{2026}
\acmYear{2026}

\setcopyright{cc}
\setcctype{by}

\acmConference[KDD '26]{Proceedings of the 32nd ACM SIGKDD Conference on Knowledge Discovery and Data Mining V.2}{August 09--13, 2026}{Jeju Island, Republic of Korea}
\acmBooktitle{Proceedings of the 32nd ACM SIGKDD Conference on Knowledge Discovery and Data Mining V.2 (KDD '26), August 09--13, 2026, Jeju Island, Republic of Korea}
\acmISBN{979-8-4007-2259-2/2026/08}

\acmDOI{10.1145/3770855.3818118}

\usepackage{microtype}
\usepackage{graphicx}
\usepackage{subcaption}
\usepackage{booktabs} 
\usepackage{algorithm}
\usepackage{algorithmic}

\usepackage{diagbox}

\usepackage{mathtools}
\usepackage{multirow}
\usepackage{xcolor}
\usepackage{enumitem}

\usepackage[capitalize,noabbrev]{cleveref}

\theoremstyle{plain}

\theoremstyle{definition}

\theoremstyle{remark}

\usepackage{xspace}

\newcommand{\method}{\textsc{CausalNeg}\xspace}
\newcommand{\loss}{\mathcal{L}}
\newcommand{\entropy}{\mathcal{H}}

\newcommand{\negset}{\mathcal{N}}

\usepackage{fontawesome5}

\begin{document}


\title{\textit{When Hard Negatives Hurt}: Bridging the Generative- Discriminative Gap in Hard Negative Synthesis for Retrieval}

\author{Zhicheng Zhang}
\authornote{These authors contributed equally to this research.}
\email{zhang-zc24@mails.tsinghua.edu.cn}
\affiliation{%
  \institution{Shenzhen International Graduate School, Tsinghua University}
  \city{Shenzhen}
  \country{China}
}

\author{Jiwei Tang}
\authornotemark[1]
\email{tangjw24@mails.tsinghua.edu.cn}
\affiliation{%
  \institution{Shenzhen International Graduate School, Tsinghua University}
  \city{Shenzhen}
  \country{China}
}

\author{Kuicai Dong}
\email{dong.kuicai@huawei.com}
\affiliation{%
  \institution{Huawei Technologies Co., Ltd.}
  \city{Shenzhen}
  \country{China}
}

\author{Xiaopeng Li}
\email{xiaopli2-c@my.cityu.edu.hk}
\affiliation{%
  \institution{City University of Hong Kong}
  \city{Hong Kong SAR}
  \country{China}
}

\author{Jieming Zhu}
\email{jiemingzhu@ieee.org}
\affiliation{%
  \institution{Huawei Technologies Co., Ltd.}
  \city{Shenzhen}
  \country{China}
}

\author{Jingyu Li}
\email{lijy768@mail2.sysu.edu.cn}
\affiliation{%
  \institution{School of Cyber Science and Technology, Sun Yat-sen University}
  \city{Guangzhou}
  \country{China}
}

\author{Qianhui Zhu}
\email{zhuqh23@mails.tsinghua.edu.cn}
\affiliation{%
  \institution{Shenzhen International Graduate School, Tsinghua University}
  \city{Shenzhen}
  \country{China}
}

\author{Fengyuan Lu}
\email{522024710011@smail.nju.edu.cn}
\affiliation{%
  \institution{School of Intelligence Science and Technology, Nanjing University}
  \city{Nanjing}
  \country{China}
}

\author{Wang Jiaheng}
\email{jwangkg@connect.ust.hk}
\affiliation{%
  \institution{The Hong Kong University of Science and Technology}
  \city{Hong Kong SAR}
  \country{China}
}

\author{Gang Wang}
\email{wanggang110@huawei.com}
\affiliation{%
  \institution{Huawei Technologies Co., Ltd.}
  \city{Shenzhen}
  \country{China}
}

\author{Hai-Tao Zheng}
\authornote{Corresponding authors.}
\email{zheng.haitao@sz.tsinghua.edu.cn}
\affiliation{%
  \institution{Shenzhen International Graduate School, Tsinghua University}
  \city{Shenzhen}
  \country{China}
}

\author{Zhaocheng Du}
\authornotemark[1]
\authornotemark[2]
\email{zhaochengdu@huawei.com}
\affiliation{%
  \institution{Huawei Noah's Ark Lab}
  \city{Shenzhen}
  \country{China}
}
\renewcommand{\shortauthors}{Zhang et al.}
\begin{abstract}

Hard negative mining has become the dominant strategy for training retrievers, yet it faces intrinsic limitations: negatives are bounded by corpus availability, selected by retriever score rather than diagnostic value, and increasingly contaminated by false positives as the retriever improves. LLM-based synthesis offers a principled alternative, where negatives that are unconstrained, targeted, and free from false positive risk.
But we show that \textit{\textbf{na\"{i}vely incorporating generated negatives into contrastive learning often degrades retrieval performance}}.
We identify and formalize the root cause as a \textbf{\textit{generative–discriminative gap}}: LLM generation optimizes for fluent, plausible text, while contrastive learning demands strategic violations of relevance at the decision boundary. 
Our analysis reveals two compounding failure modes: \textit{discriminative-agnostic generation}, where the LLM lacks an explicit model of query information needs and defaults to generic or topic-drifted text that provides no contrastive signal; and \textit{source-dependent shortcuts}, where distributional artifacts enable the model to distinguish negatives by origin rather than relevance, causing gradient drift that actively corrupts optimization.
To close this gap, we propose \method consisting of two main modules:
\textbf{\textit{(1) CoT-guided counterfactual perturbation for data construction}}: decomposes why a document satisfies a query into explicit information requirements, then surgically violates individual requirements to construct negatives with controlled, interpretable hardness. 
\textbf{\textit{(2) Query-view entropy maximization during training:}} disperses generated negatives across the similarity spectrum, minimizing the mutual information between source identity and similarity scores to suppress shortcut exploitation.
Experiments on 4 retrieval benchmarks show that \method outperforms mining-only and na\"{i}ve generation baselines, validating causally grounded synthesis and entropy-regularized training as complementary solutions to the generative–discriminative gap.
\end{abstract}





\maketitle

\newcommand\kddavailabilityurl{https://doi.org/10.5281/zenodo.20404602}
\ifdefempty{\kddavailabilityurl}{}{
\begingroup\small\noindent\raggedright\textbf{Resource Availability:}\\
The source code of this paper has been made publicly available at
\url{\kddavailabilityurl}. The code repository is available at
\url{https://github.com/mzhangzhicheng/CausalNeg}.
\endgroup
}

\section{Introduction}
\label{sec:intro}

\begin{figure*}[t]
    \centering
    \includegraphics[width=\textwidth]{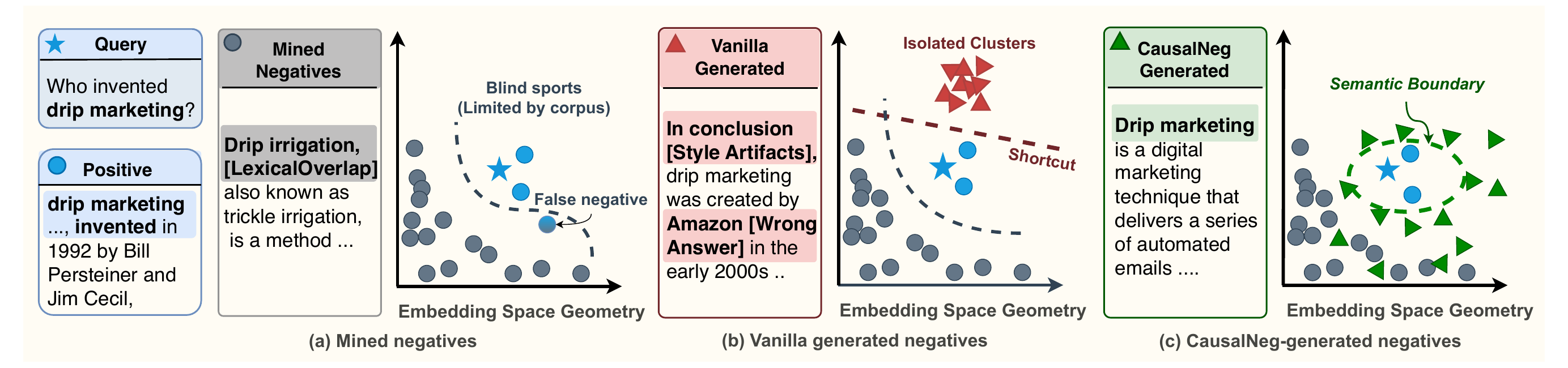}
    \vspace{-2.2em}
    \caption{Illustrative comparison of negative types. (a) Mined negatives are constrained by corpus availability and suffer from blind spots and false-negative risk. (b) Vanilla generated negatives carry style artifacts and factual errors, creating source-dependent shortcuts. (c) \method-generated negatives violate specific information requirements while preserving topical coherence, yielding well-distributed samples that refine semantic boundaries.}
    \label{fig:intro}
\end{figure*}

Information retrieval (IR) seeks to identify relevant documents from a large corpus given a user query~\cite{yates-etal-2021-pretrained}. Modern retrievers~\cite{karpukhin-etal-2020-dense} are typically initialized from pre-trained LLMs~\cite{qwen3embedding,lee2025geminiembeddinggeneralizableembeddings}, and optimized via contrastive learning that teaches the model to distinguish relevant documents from irrelevant ones. Such discriminative objective~\cite{oord2019representationlearningcontrastivepredictive, 10.5555/3524938.3525087} relies heavily on the quality of positive and negative training pairs. 
In particular, hard negatives (documents that appear superficially relevant to a query yet fail to satisfy its actual information need) are essential for shaping a fine-grained decision boundary. Without sufficiently challenging negatives, the contrastive loss provides near-zero gradient~\cite{TriSampler}, and the model plateaus at a coarse level of discrimination~\cite{zhou-etal-2022-simans}. Hard negative mining~\cite{xiong2021approximate,10.1145/3404835.3462880}, which retrieves high-scoring but irrelevant passages from the corpus using an existing retriever, has become the dominant strategy for constructing such training signals and has driven substantial gains across retrieval benchmarks~\cite{thakur2021beir,marco}.

Despite its effectiveness, hard negative mining has several intrinsic limitations.
\textit{\textbf{(1) lack of diversity}.}
They are bounded by what exists in the corpus~\cite{TriSampler,inpars}: if no passage exhibits a particular failure mode (e.g., subtle factual deviation, entity confusion, or logical inversion), it simply cannot be mined.
\textit{\textbf{(2) Uncontrolled intentionality.}} 
Even among negatives that do exist, mining selects them by retriever score rather than by diagnostic value. A passage is surfaced because it scores high, not because it tests a specific aspect of the query's information need. This leads to accidental hardness (e.g., lexical overlap or topical proximity) rather than targeted at fine-grained semantic or factual discrimination.
\textit{\textbf{(3) False negative risk.}} As the retriever improves during iterative mining, the highest-scoring negatives increasingly tend to be unlabeled positives that genuinely satisfy the query. Training against these mislabeled examples pushes relevant documents away from the query, actively corrupting the decision boundary~\cite{qu-etal-2021-rocketqa}.
Collectively, these limitations impose a quality ceiling: mining alone cannot fully exploit the representational capacity of modern LLM-based retrievers.

A natural alternative is to leverage LLMs to \emph{generate} hard negatives on demand. In principle, synthesis offers a compelling solution to mining's limitations: generated negatives are unconstrained by corpus availability, can be designed to target specific failure modes, and avoid false negative contamination. However, our study (\S\ref{sec:analysis}) reveals that na\"{i}vely incorporating LLM-generated negatives into contrastive learning often fails, and may even degrade retrieval performance. The core issue lies in a fundamental mismatch: \textit{LLM generation is a generative process optimized for producing fluent, plausible text, while contrastive learning is a discriminative task that requires maximizing separation at the relevance decision boundary.} This \textbf{\textit{generative-discriminative gap}} manifests in two critical challenges:
\textit{\textbf{(1) Discriminative-agnostic generation}} (\S\ref{ssec:discriminative_agnostic}). The LLM lacks explicit knowledge of what constitutes a ``negative'' in the retrieval sense. It generates plausible text, not strategically non-responsive text. Vanilla prompting thus produces samples that violate contrastive semantics: overly generic descriptions, topic-drifted passages, or incorrect query answers. Such samples fail to provide meaningful training signals and exhibit pathological embedding behaviors such as isolated clusters and diversity collapse.
\textit{\textbf{(2) Source-dependent shortcuts}} (\S\ref{ssec:source_shortcuts}). The generated negatives (even semantically satisfying the discriminative requirements) carry a distributional signature distinct from real corpus data. This source-specific fingerprint enables the model to discriminate negatives by their stylistic artifacts rather than by relevance, creating a shortcut that bypasses genuine semantic learning. We show that exploiting such shortcuts causes gradient drift, pulling optimization away from the retrieval objective toward an artifact-biased solution.

To address these challenges, we propose \textbf{\method} (Causal Negative Synthesis), a unified framework that leverages causal reasoning to systematically construct hard negatives through counterfactual perturbation of query information requirements.
Our framework introduces two synergistic components. First, \textbf{\textit{Chain-of-Thought Guided Counterfactual Perturbation}} reconceptualizes negative generation as structured reasoning: given query $q$ and positive document $d^+$, we decompose \emph{why} $d^+$ satisfies $q$'s information requirements into a reasoning chain, then systematically perturb critical nodes to construct negatives that violate specific requirements. This transforms generation from discriminative-agnostic text synthesis into targeted construction of interpretable failure modes with controlled diagnostic value. Second, \textbf{\textit{Query-View Entropy Maximization (QEM)}} addresses source-dependent shortcuts through a theoretically grounded training intervention. We maximize the entropy of generated negatives' query-conditioned similarity distribution while enforcing probability balance across sources, thereby minimizing the mutual information between source identity and similarity scores. Critically, QEM backpropagates only through generated embeddings, using mined negatives as a fixed reference distribution to harmonize both sources without corrupting the mining signal. Together, these components enable a paradigm shift from mining-based \emph{discovery} of accidental hardness to synthesis-based \emph{design} of intentional hardness, while maintaining corpus distribution alignment.

Our main contributions are summarized as follows:
\begin{itemize}[leftmargin=*, itemsep=0em, topsep=-0.1em]

\item We identify and formalize the \emph{generative-discriminative gap} in LLM-based negative synthesis, characterizing two failure modes: discriminative-agnostic generation and source-dependent shortcuts, and their impact on gradient dynamics through mutual information analysis.

\item We introduce \emph{Chain-of-Thought Guided Counterfactual Perturbation}, which reconceptualizes negative generation as causal reasoning: decomposing relevance relationships and perturbing critical nodes to synthesize interpretable hard negatives, shifting from accidental to intentional hardness.

\item We propose \emph{Query-View Entropy Maximization (QEM)}, a training intervention that suppresses shortcuts by minimizing mutual information through entropy maximization and probability balance, while preserving signals from both negative sources.

\item Extensive experiments show \method significantly outperforms mining-only and na\"{i}ve generation baselines, validating causally grounded synthesis and entropy-regularized training.

\end{itemize}

\section{Why Synthetic Negatives Fail: Diagnosing the Generative-Discriminative Gap}
\label{sec:analysis}

The introduction (\S\ref{sec:intro}) identifies a \emph{generative-discriminative gap} that manifests as two challenges: discriminative-agnostic generation and source-dependent shortcuts. Before presenting our method, we empirically ground both claims. We first show that na\"{i}ively generating and mixing LLM-generated negatives degrades retrieval performance (\S\ref{ssec:performance_pilot}), then diagnose why: generated negatives exhibit deficiencies that reflect discriminative-agnostic generation (\S\ref{ssec:discriminative_agnostic}), and even when semantically adequate, they introduce source-dependent shortcuts that corrupt optimization dynamics (\S\ref{ssec:source_shortcuts}).

\subsection{Analysis Setup}
\label{sec:analysis_setup}
We conduct all analyses on mMARCO-zh~\cite{Bonifacio2021MMarco,marco} dataset. We randomly sample 2,000 queries from the training set and construct, for each query, a tuple of four elements: the query itself, one positive document, fifteen mined hard negatives, and three vanilla-generated hard negatives. Mined hard negatives are the highest-scoring non-positive BM25 result from the full document corpus. Vanilla generations are hard negatives produced by prompting a strong proprietary instruction-tuned LLM with the query and positive document using the SyNeg~\cite{li2024synegllmdrivensynthetichardnegatives} generation strategy, which represents a standard vanilla approach without structured reasoning or explicit relevance modeling. This controlled setup ensures that any distributional divergence we observe is attributable to the generation process itself rather than confounds such as topic distribution or quantity imbalance.
For embedding-space analyses (\S\ref{ssec:discriminative_agnostic}), we encode all documents with a pre-trained Qwen3-Embedding-0.6B encoder~\cite{qwen3embedding}. 
For training dynamics (\S\ref{ssec:source_shortcuts}), we fine-tune Qwen3-0.6B with a standard InfoNCE objective, mixing mined and generated negatives.
Please refer to more training details in Appendix \ref{app:experiments}.

\subsection{Na\"{i}ve Synthesizing and Mixing Negatives Degrade Retrieval Performance}
\label{ssec:performance_pilot}
As a motivating observation, we compare three training configurations: (i) mined negatives only, (ii) vanilla-generated negatives, and (iii) a na\"{i}ve mixture of both. Table~\ref{tab:pilot} shows that generated-only training yields near-zero NDCG@10 on most benchmarks. Na\"{i}ve mixing offers a marginal gain on mMARCO-zh but hurts on HotpotQA, NQ, and TQA. This confirms that generated hard negatives, despite being designed to be challenging, introduce pathologies that offset their intended benefit. The remainder of this section diagnoses the root causes.

\begin{table}[t]
    \centering
    \setlength{\tabcolsep}{4pt}
    \caption{Pilot study (NDCG@10). Vanilla-generated negatives training fails catastrophically; na\"{i}ve mixing degrades 3 of 4 datasets.}
    \vspace{-1em}
    \label{tab:pilot}
    \begin{tabular}{lcccc}
        \toprule
        Configuration & mMARCO-zh & HotpotQA & NQ & TQA \\
        \midrule
        Mined-only     & 68.82 & 55.88 & 77.15 & 56.75 \\
        Vanilla Generation & 0.00  & 6.82  & 0.02  & 0.00  \\
        Na\"{i}ve Mixture  & 69.31 & 55.39 & 74.33 & 53.16 \\
        \midrule
        \method (\textit{ours})  & \textbf{71.78} & \textbf{57.23} & \textbf{78.17} & \textbf{59.02} \\
        \bottomrule
    \end{tabular}
\end{table}

\subsection{Discriminative-Agnostic Generation}
\label{ssec:discriminative_agnostic}
We first ask: \emph{what properties of generated negatives make them ineffective for contrastive learning?} We combine embedding-space geometry with LLM-based semantic diagnosis to show that the generation process produces negatives that are distributionally and semantically misaligned with discriminative training.

\begin{figure}[t]
    \centering
    \begin{subfigure}{0.495\columnwidth}
        \centering
        \includegraphics[width=\linewidth]{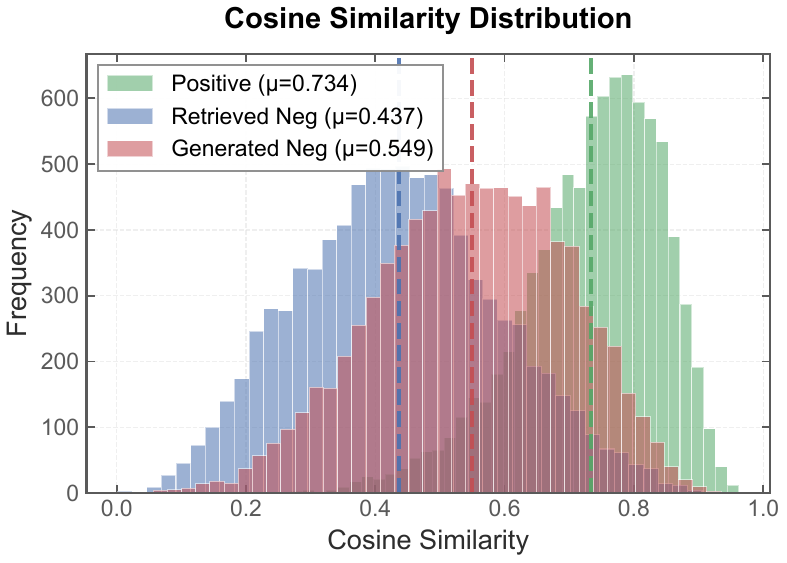}
        \vspace{-1.8em}
                \caption{Similarity distribution}
                        \label{fig:naive_similarity}

    \end{subfigure}
    \hfill
    \begin{subfigure}{0.495\columnwidth}
        \centering
        \includegraphics[width=\linewidth]{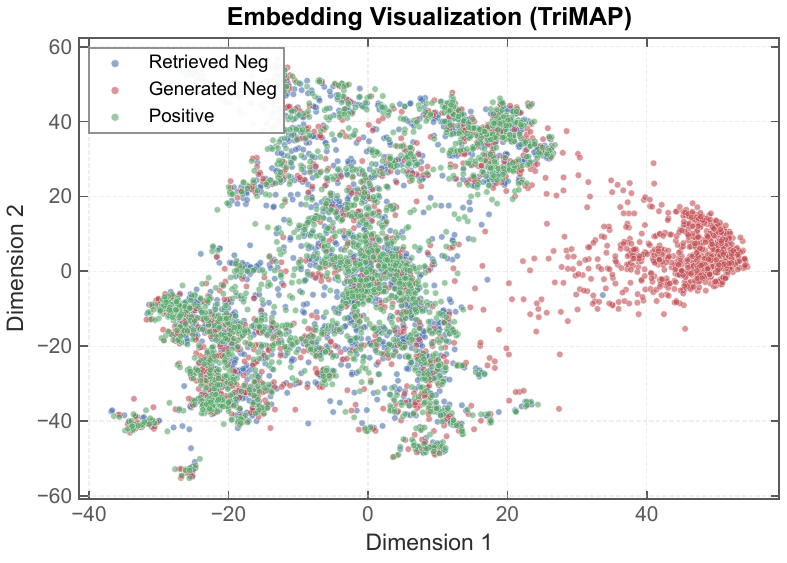}
        \vspace{-1.8em}
                \caption{TriMAP visualization}
                        \label{fig:naive_clustering}

    \end{subfigure}
    \vspace{-2em}
    \caption{Analysis of generated negatives: (a) cosine similarity distribution between queries and documents; (b) TriMAP visualization with cluster assignments.}
    \label{fig:embedding_analysis}
\end{figure}

\paragraph{\textbf{Distributional divergence in embedding space}.}
Cosine similarity between queries and different document types reveals a consistent ordering: positives exhibit the highest similarity, followed by generated negatives, then mined negatives (Figure~\ref{fig:embedding_analysis}a).
On the surface, higher similarity might suggest greater hardness. However, the distributional \emph{shape} of generated negatives differs qualitatively from that of mined negatives, indicating that their proximity to queries arises from different underlying features: superficial plausibility rather than targeted near-miss relevance.

Clustering analysis sharpens this observation. Applying HDBSCAN~\cite{campello2013density} to the combined embeddings of all document types, we find that queries, positives, and mined negatives distribute across clusters without forming type-exclusive groups.
In contrast, approximately 24\% of generated negatives aggregate into \emph{pure clusters} that contain no other document type (Figure~\ref{fig:embedding_analysis}b). No such source-exclusive clustering occurs for queries, positives, or mined negatives.
Moreover, this structure is \emph{query-agnostic}: across three independent generation runs, the set of queries whose negatives fall into pure clusters varies substantially (union: 1,085; intersection: 281), confirming that pure cluster formation is a stochastic artifact of the generation process rather than a property of specific queries.

\paragraph{\textbf{Semantic causes of distributional anomalies}.}
The geometric evidence reveals that generated negatives diverge, but not why. To obtain interpretable explanations, we compare generated negatives that cluster normally with other document types (``good'' samples) against those forming pure clusters (``bad'' samples) using an LLM-based diagnostic framework. Following prior work on describing distributional differences via natural language hypotheses~\cite{pmlr-v162-zhong22a, singh-etal-2023-explaining,swayamdipta-etal-2020-dataset}, we discover and validate failure-mode hypotheses, then annotate 600 samples across eight semantic dimensions. Full protocol details are provided in Appendix~\ref{app:llm_diagnosis}.

\begin{table}[t]
    \centering
    \caption{Failure modes distinguishing bad generated negatives (pure clusters) from good ones (mixed clusters). $\Delta$: prevalence difference (bad $-$ good).}
    \vspace{-1em}
    \label{tab:failure_modes}
    \begin{tabular}{lr}
        \toprule
        Failure Mode & $\Delta$ (\%) \\
        \midrule
        F1: Content over-generalization & +25.0 \\
        F2: Domain mismatch / topic drift & +14.0 \\
        F3: Semantic isolation / noise & +12.8 \\
        F4: Core information avoidance & +11.2 \\
        F5: Background-only (vs.\ wrong entity) & +11.0 / $-$15.0 \\
        F6: Medium specificity (vs.\ very specific) & +13.3 / $-$14.0 \\
        \bottomrule
    \end{tabular}
\end{table}

Table~\ref{tab:failure_modes} reveals six failure modes with a coherent structure. F1 through F4 share a common cause: the LLM fails to engage with the query's specific information need, manifesting as generic background text (F1), topic drift (F2), semantic noise (F3), or deliberate vagueness (F4).
F5 is the most revealing. Good samples invalidate relevance through specific factual substitution (e.g., wrong entity or wrong time) while preserving topical proximity; bad samples instead default to vague background content bearing little resemblance to the positive document. This contrast exposes the core deficiency: effective negatives require reasoning about why the positive satisfies the query and surgically perturbing that relationship, which vanilla generation fundamentally lacks. F6 reinforces this from a stylistic angle, with bad samples exhibiting pronounced template patterns and anomalous uniformity.

\subsection{Source-Dependent Shortcuts}
\label{ssec:source_shortcuts}

The preceding analysis shows that a substantial fraction of generated negatives are semantically deficient. A natural follow-up question is: \emph{if we fix the semantic quality, does the problem disappear?} It does not. Even semantically adequate generated negatives carry stylistic and structural artifacts of LLM generation that contrastive learners exploit as shortcuts in lieu of genuine relevance discrimination.

\begin{figure}[t]
    \centering
    \includegraphics[width=\columnwidth]{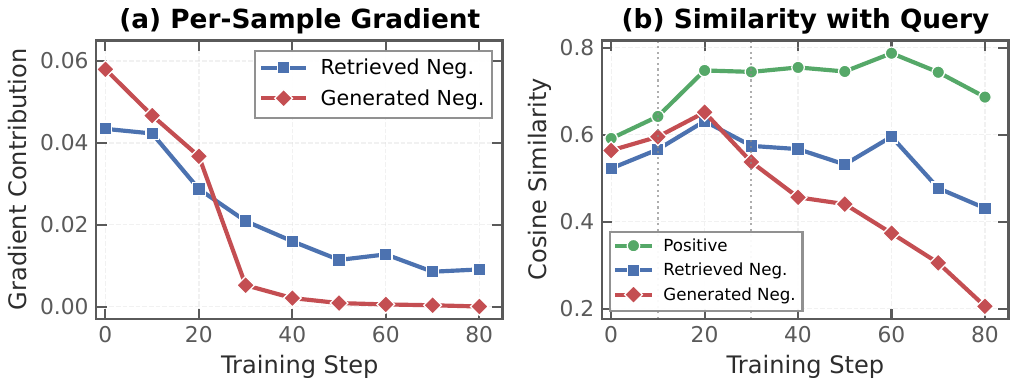}
    \vspace{-2.5em}
    \caption{Training dynamics under mixed negatives. (a) Per-sample gradient weight for mined and generated negatives. (b) Cosine similarity between queries and each document type. Generated negatives are rapidly ``solved,'' and their gradient contribution collapses, while mined negatives remain hard throughout training.}
    \label{fig:dynamics}
\end{figure}

\paragraph{\textbf{Stage transition in learning signal}.}
We track the mean softmax probability of each negative type in the InfoNCE denominator (i.e., the per-sample gradient weight) and the cosine similarity between queries and each document type across training checkpoints.
Figure~\ref{fig:dynamics} reveals a clear three-phase transition. In the \emph{early phase}, generated negatives receive comparable or larger gradient weight than mined negatives, consistent with their higher initial query similarity. In the \emph{transition phase}, the model discovers superficial cues that distinguish generated from mined negatives, and the generated-negative contribution begins to collapse. In the \emph{late phase}, this collapse becomes dramatic: the generated-negative weight drops by over three orders of magnitude while the mined-negative weight remains substantial.
Mined negatives become the sole effective source of learning pressure, rendering the generated negatives inert.

\paragraph{\textbf{Gradient drift: misalignment, not vanishing}.}
If the softmax collapse simply meant that generated negatives ceased to matter, the situation would be benign, equivalent to training on mined negatives alone. Table~\ref{tab:gradient_drift_detailed} reveals otherwise. As training progresses, the loss contribution of generated negatives becomes negligible, yet their gradient magnitude does not shrink; the gradient norm ratio (generated over mined) in fact \emph{increases}. 
Meanwhile, the cosine similarity between positive and generated-negative gradients weakens from 0.73 to 0.24, indicating that these gradients drift away from relevance discrimination and toward source-identity discrimination. The result is not a vanishing signal but a persistent, misaligned one: substantial gradient magnitude directed along a spurious axis.

\begin{table}[t]
\caption{Gradient dynamics across training stages. Generated negatives exhibit \textit{gradient drift}: their loss contribution collapses while gradient magnitude increases and direction becomes less antagonistic to the positive gradient.}
\vspace{-1.5em}
\label{tab:gradient_drift_detailed}
\vskip 0.1in
\centering
\small
\begin{tabular}{lcccc}
\toprule
\textbf{Stage} & \textbf{Contrib.} & \textbf{Contrib.} & \textbf{Alignment} & \textbf{Magnitude} \\
 & \textbf{(Gen)} & \textbf{Ratio} & \textbf{(Pos-Gen)} & \textbf{Ratio} \\
\midrule
\multirow{2}{*}{Early (0/10)}
 & 0.058 & 1.34 & --- & --- \\
 & 0.047 & 1.10 & $-0.73$ & 0.91 \\
\midrule
\multirow{2}{*}{Mid (20/30)}
 & 0.037 & 1.28 & $-0.65$ & 1.02 \\
 & 0.005 & 0.25 & $-0.53$ & 1.22 \\
\midrule
\multirow{2}{*}{Late (50/80)}
 & 0.001 & 0.08 & $-0.42$ & 1.32 \\
 & $4.14\times 10^{-5}$ & $4.6\times 10^{-3}$ & $-0.24$ & 1.42 \\
\bottomrule
\end{tabular}
\vskip 0.05in
\raggedright
\footnotesize
\textbf{Contrib.}: Per-sample softmax probability (gradient weight).
\textbf{Contrib.\ Ratio}: Generated / Mined.
\textbf{Alignment}: Gradient cosine similarity with positive gradient (negative values indicate opposing directions).
\textbf{Magnitude Ratio}: Gradient norm ratio (Generated / Mined).
\end{table}

This pattern constitutes direct evidence for the source-dependent shortcut: the latent source variable $Z$ induces an auxiliary optimization objective that competes with relevance discrimination. The model learns to ``solve'' generated negatives by detecting their distributional fingerprint, producing substantial gradient updates but misaligned with the true decision boundary. The result is not harmless redundancy but active interference with the retrieval objective.

\section{Method: \method}
\label{sec:method}

\begin{figure*}[t]
    \centering
    \includegraphics[width=0.92\textwidth]{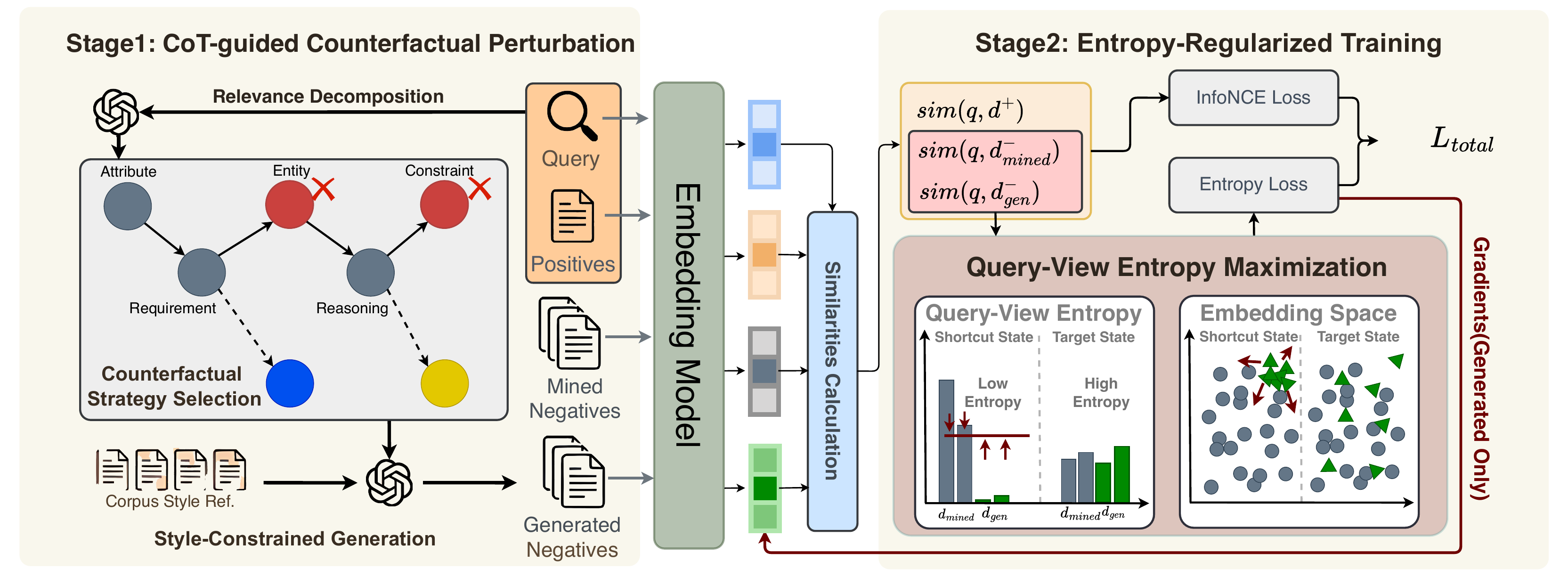}
    \vspace{-1.0em}
    \caption{Overview of the \method. (Left) CoT-guided generation decomposes query into a conjunction of information requirements and applies counterfactual perturbations to produce diverse hard negatives. (Right) Entropy-regularized training maximizes the entropy of generated negatives under the query-conditioned similarity distribution, suppressing source-dependent shortcuts. Gradients from the entropy loss propagate only through generated negative embeddings.}
    \label{fig:pipeline}
\end{figure*}

The two pathologies (discriminative-agnostic generation and source-dependent shortcuts) diagnosed in Section~\ref{sec:analysis} operate at different stages of the training pipeline: one at data construction, the other at learning time, and therefore demand complementary interventions. We introduce \method, a framework with two components: \emph{CoT-guided Counterfactual Perturbation} (\S\ref{sec:generation}), which replaces free-form synthesis with structured reasoning over query–document relevance to produce interpretable hard negatives, and \emph{Query-View Entropy Maximization} (\S\ref{sec:entropy}), a training-time regularizer that suppresses source-dependent shortcuts by minimizing the mutual information between source identity and similarity scores. We begin with the shared problem formalization (\S\ref{ssec:formalization}).

\subsection{Problem Formalization}
\label{ssec:formalization}

Let $q$ denote a query, $d^+$ a relevant (positive) document, and $\negset = \negset_{\text{mined}} \cup \negset_{\text{gen}}$ the set of negatives comprising mined and generated subsets. Standard contrastive training minimizes the InfoNCE loss:
\begin{equation}
    \loss_{\text{InfoNCE}} = -\log \frac{\exp(s(q, d^+)/\tau)}{\exp(s(q, d^+)/\tau) + \sum_{d^- \in \negset} \exp(s(q, d^-)/\tau)},\label{eq:infonce}
\end{equation} 
where $s(\cdot, \cdot)$ denotes cosine similarity of embeddings and $\tau$ is a temperature parameter.
As shown in Section~\ref{ssec:source_shortcuts}, the mixed negative pool introduces a latent source variable $Z \in \{\text{mined}, \text{gen}\}$.
When the model can predict $Z$ from the similarity score $s(q, d^-)$, it exploits this shortcut rather than learning genuine relevance discrimination. Formally, shortcut learning occurs when the mutual information $I(Z;\; s(q, d^-)) > 0$. Our framework targets both the \emph{cause} of this mutual information (distributional divergence introduced at generation time) and its \emph{exploitation} (shortcut learning at training time).

\subsection{CoT-guided Counterfactual Perturbation}
\label{sec:generation}

\paragraph{\textbf{Core principle: relevance as a conjunction of satisfiable requirements}.}

The failure modes in Table~\ref{tab:failure_modes} reveal that effective negatives invalidate relevance through precise factual substitution, while poor ones default to generic background. This points to a structural insight: relevance is not a monolithic judgment but a conjunction of \emph{information requirements} $\mathcal{R}(q) = \{r_1, r_2, \ldots, r_K\}$ that the query implicitly demands:
\begin{equation}
    \text{Rel}(q, d) = 1 \iff \wedge_{k=1}^{K} \phi_k(d, r_k) = 1,
    \label{eq:relevance_conjunction}
\end{equation}
where $\phi_k(d, r_k)$ indicates whether $d$ satisfies the $k$-th requirement. A \emph{counterfactual negative} $d^-$ violates exactly one requirement while preserving all others:
\begin{equation}
    \exists!\; j: \quad \phi_j(d^-, r_j) = 0 \;\;\wedge\;\; \forall\, k \neq j,\; \phi_k(d^-, r_k) = 1.
    \label{eq:counterfactual}
\end{equation}
This single-requirement violation (i) guarantees the \emph{hardness} ($K{-}1$ of $K$ requirements satisfied, maintaining topical proximity) and (ii) makes \emph{the violated requirement identifiable}.
It also explains why vanilla generation fails: without an explicit model of $\mathcal{R}(q)$, the LLM violates requirements in an uncontrolled fashion.

\paragraph{\textbf{From principle to pipeline: CoT-guided decomposition and perturbation}.}
Achieving Equation~\ref{eq:counterfactual} requires three capabilities: (i) extracting the requirement set $\mathcal{R}(q)$ from a query-document pair, (ii) selecting which requirement to violate and how, and (iii) generating a document that realizes the perturbation while remaining naturalistic. We implement these through a structured LLM pipeline inspired by
chain-of-thought prompting~\cite{wei2023chainofthoughtpromptingelicitsreasoning,jia-etal-2025-bridging},
decomposed into three stages.

\textit{Stage 1: Relevance decomposition.}
Given query $q$ and positive document $d^+$, we prompt the LLM to produce a chain-of-thought that decomposes \emph{why} $d^+$ satisfies $q$ into a structured set of $K$ information requirements $\{r_1, \ldots, r_K\}$. Each requirement is a concrete, falsifiable condition (e.g., ``\textit{identifies the specific enzyme responsible for X}'' or ``\textit{provides a temporal ordering consistent with the 2019 timeline}''). This stage makes the implicit conjunction in Equation~\ref{eq:relevance_conjunction} explicit, providing the scaffold for targeted perturbation.

\textit{Stage 2: Counterfactual strategy selection.}
For each requirement $r_k$, we design perturbation strategies that produce a violation $\phi_k(d^-, r_k) = 0$ while preserving topical coherence. Three canonical perturbation types cover the space of meaningful requirement violations: \emph{entity substitution} replaces the target entity with a related but incorrect alternative (e.g., substituting a sibling enzyme), invalidating the identification requirement; \emph{requirement drift} shifts the document's focus to an adjacent but distinct information need (e.g., describing the enzyme's structure rather than its function), breaking the functional requirement; and \emph{constraint violation} breaks specific boundary conditions (temporal, spatial, causal, or quantitative) that the query demands (e.g., describing events from 2017 instead of 2019). Multiple strategies per requirement ensure diversity across generated negatives, preventing the model from learning to recognize a single perturbation pattern.

\textit{Stage 3: Style-controlled generation.}
Each perturbation strategy, combined with reference documents sampled from the corpus, is provided to the LLM for generation. The style references serve a dual purpose: they constrain the surface form to match the corpus distribution (mitigating the stylistic artifacts identified in Table~\ref{tab:failure_modes}), and they anchor the generated text in target domain. Each generated negative is accompanied by a trace specifying which requirement it violates and through which perturbation type, enabling interpretability and downstream quality filtering. Complete prompt templates are provided in Appendix~\ref{app:prompts}.

\subsection{Query-View Entropy Maximization}
\label{sec:entropy}

\paragraph{\textbf{Connecting to the diagnosed failure.}}
The three-phase collapse in Section~\ref{ssec:source_shortcuts} occurs because generated negatives cluster at characteristic similarity values, enabling the model to predict source identity from $s(q, d^-)$ alone. Once this shortcut is discovered, generated-negative gradient weight drops and drifts away from relevance discrimination (Table~\ref{tab:gradient_drift_detailed}). Our regularizer targets the root cause directly: by preventing similarity clustering, it blocks the transition phase that triggers the collapse, maintaining generated negatives as effective training signals throughout optimization.

\paragraph{\textbf{Theoretical Grounding}.}

Let $Z \in \{\text{mined}, \text{gen}\}$ denote the source indicator. The mutual information between source identity and similarity score decomposes as:
\begin{equation}
    I(Z;\; s(q, d^-)) = H(Z) - H(Z \mid s(q, d^-)).
\end{equation}
Since the source prior $H(Z)$ is fixed by the data mixture, suppressing this mutual information reduces to increasing $H(Z \mid s(q, d^-))$, i.e., making source identity unpredictable from the similarity score. When generated negatives concentrate at characteristic similarity values, a simple threshold on $s(q, d^-)$ predicts $Z$ with high accuracy, yielding low $H(Z \mid s)$. Dispersing generated negatives across the same similarity range as mined negatives eliminates this diagnostic signal and pushes $H(Z \mid s)$ toward its maximum. This directly motivates maximizing the entropy of generated negatives' query-conditioned similarity distribution.

\paragraph{\textbf{Entropy Maximization Objective}.}
For each query $q$, we compute scaled similarities to all negatives and convert them to a probability distribution:
\begin{equation}
    p_i = \frac{\exp(s(q, d_i^-) / \tau_{\text{ent}})}{\sum_j \exp(s(q, d_j^-) / \tau_{\text{ent}})}, \quad d_i^- \in \negset_{\text{mined}} \cup \negset_{\text{gen}}.
\end{equation}
The auxiliary loss comprises two complementary terms. The first maximizes the entropy of the generated negatives' probability distribution, preventing clustering at particular similarity values:
\begin{equation}
    \entropy_{\text{gen}} = -\sum_{d^- \in \negset_{\text{gen}}} \tilde{p}_{d^-} \log \tilde{p}_{d^-},
\end{equation}
where $\tilde{p}$ denotes probabilities renormalized over generated negatives only. The second enforces balance between the aggregate probability masses of the two negative sources, preventing the model from pushing all generated negatives to extreme similarity values (which would technically maximize within-group entropy but eliminate their training utility):
\begin{equation}
    \loss_{\text{balance}} = \left( \sum_{d^- \in \negset_{\text{gen}}} p_{d^-} - \frac{|\negset_{\text{gen}}|}{|\negset_{\text{mined}}| + |\negset_{\text{gen}}|} \right)^2.
\end{equation}
The combined auxiliary loss is: $\loss_{\text{entropy}} = -\entropy_{\text{gen}} + \loss_{\text{balance}}$.

\paragraph{\textbf{Selective gradient propagation}.}
The final training objective combines InfoNCE and entropy loss with regularization factor $\lambda$:
\begin{equation}
    \loss_{\text{total}} = \loss_{\text{InfoNCE}} + \lambda \cdot \loss_{\text{entropy}},
\end{equation}

A critical design choice is that the entropy loss backpropagates only through the generated negative embeddings; both mined negative embeddings and the query embedding are detached from the entropy computation graph. This asymmetric gradient design serves two purposes. First, it preserves the learning signal from mined negatives: since mined negatives come from the real corpus and are unaffected by the generative-discriminative gap, their contribution to InfoNCE should remain unperturbed. Second, it prevents the query encoder from being pulled toward a distribution-matching objective that is orthogonal to relevance discrimination. The entropy loss thus acts as a targeted correction that reshapes how generated negatives are positioned in the similarity space, without distorting the core contrastive objective.

\begin{algorithm}[t]
    \caption{Query-View Entropy Loss Computation}
    \label{alg:entropy}
    \begin{algorithmic}[1]
        \REQUIRE Query embeddings $q$ [B, D], mined negatives $\negset_m$ [B, $N_m$, D], generated negatives $\negset_g$ [B, $N_g$, D], temperature $\tau$
        \STATE Detach query and mined negatives: $q \leftarrow \text{detach}(q)$, $\negset_m \leftarrow \text{detach}(\negset_m)$
        \STATE Concatenate: $\negset_{\text{all}} \leftarrow [\negset_m; \negset_g]$ \COMMENT{[B, $N_m$+$N_g$, D]}
        \STATE Compute similarities: $s \leftarrow q \cdot \negset_{\text{all}}^\top / \tau$ \COMMENT{with $q$ detached}
        \STATE Softmax: $p \leftarrow \text{softmax}(s)$ \COMMENT{[B, $N_m$+$N_g$]}
        \STATE Extract generated probs: $p_g \leftarrow p[:, N_m:]$
        \STATE Normalize: $\tilde{p}_g \leftarrow p_g / \sum p_g$
        \STATE Entropy: $\entropy \leftarrow -\sum \tilde{p}_g \log \tilde{p}_g$
        \STATE Expected prob: $p_{\text{exp}} \leftarrow N_g / (N_m + N_g)$
        \STATE Balance: $\loss_{\text{bal}} \leftarrow (\sum p_g - p_{\text{exp}})^2$
        \RETURN $-\entropy + \loss_{\text{bal}}$
    \end{algorithmic}
\end{algorithm}

\section{Experiments}
\label{sec:experiments}

\begin{table*}[t]
    \centering
    \caption{Main retrieval results. \method consistently achieves the best or competitive performance. Generated-only training fails catastrophically, confirming the severity of identified failure modes. Bold: best; underline: second best.}
    \vspace{-1.0em}
    \label{tab:main_results}
    \begin{tabular}{l|cccccccc|cc}
        \toprule
        \multirow{2}{*}{\backslashbox{\textbf{Method}}{\textbf{Metric}}}
        
        & \multicolumn{2}{c}{\textbf{mMARCO-zh}} & \multicolumn{2}{c}{\textbf{HotpotQA}} & \multicolumn{2}{c}{\textbf{NQ}} & \multicolumn{2}{c}{\textbf{TQA}} & \multicolumn{2}{|c}{\textbf{Average}} \\

        & NDCG & Recall 
        & NDCG & Recall
        & NDCG & Recall
        & NDCG & Recall
        & NDCG & Recall \\
        
        \midrule
        Random 
        & 66.48 & 80.41 
        & 54.49 & \textbf{57.05} 
        & 76.70 & \underline{88.63} 
        & 53.56 & \underline{64.78}
        & 62.81 & \underline{72.72} \\

        Mined-only 
        & 68.82 & 81.27 
        & \underline{55.88} & 54.25 
        & 77.15 & 86.74 
        & \underline{56.75} & 64.54 
        & \underline{64.65} & 71.70 \\

        Vanilla Generation
       & 0.00 & 0.00
        & 6.82 & 9.23 
        & 0.02 & 0.03 
        & 0.00 & 0.00
        & 1.71 & 2.32 \\
                
        CoT Generation 
        & 34.58 & 45.34 
        & 11.17 & 11.60 
        & 0.05 & 0.10 
        & 0.00 & 0.00 
        & 11.45 & 14.26 \\

        Na\"{i}ve Mixture (SyNeg)
        & \underline{69.31} & \underline{82.00} 
        & 55.39 & 53.70 
        & 74.33 & 84.15 
        & 53.16 & 61.14 
        & 63.05 & 70.25 \\

        \midrule
        
        \textbf{\method} 
        & \textbf{71.78} & \textbf{84.05} 
        & \textbf{57.23} & \underline{55.35} 
        & \textbf{78.17} & \textbf{89.85} 
        & \textbf{59.02} & \textbf{67.11} 
        & \textbf{66.55} & \textbf{74.09} \\
        \bottomrule
    \end{tabular}
\end{table*}

\subsection{Experimental Setup}
\label{sec:setup}

\paragraph{\textbf{Datasets and Evaluation Metrics}.}
We conduct experiments on 4 retrieval benchmarks: \textbf{mMARCO-zh}~\cite{Bonifacio2021MMarco} (Chinese translation of MS~MARCO \cite{marco}), \textbf{HotpotQA}~\cite{yang-etal-2018-hotpotqa}, \textbf{NQ}~\cite{kwiatkowski-etal-2019-natural}, and \textbf{TriviaQA}~\cite{joshi-etal-2017-triviaqa}. 
We follow \citet{chen-etal-2024-m3} for the training set\footnote{\url{https://huggingface.co/datasets/Shitao/bge-m3-data}}, and \citet{qwen3embedding}  for evaluation set and metrics\footnote{\url{https://github.com/QwenLM/Qwen3-Embedding/tree/main/evaluation}}.
Specifically, we report NDCG@10 and Recall@10 as primary metrics.
Refer to more dataset and evaluation details in Appendix~\ref{app:datasets}.

\paragraph{\textbf{Implementation}.}
We select Qwen3-0.6B~\cite{qwen3embedding} as backbone and fine-tune it using the Swift framework~\cite{swift}, with approximately 10K training queries per dataset. For \method, CoT-guided negatives are generated using a strong proprietary LLM API (typically 4--10 per query); the main results use 3 per query for fair comparison with vanilla generation~\cite{li2024synegllmdrivensynthetichardnegatives}. All experiments are conducted on 8 consumer-grade 24GB accelerators with full-parameter training. Complete hyperparameters are in Appendix~\ref{app:experiments}.

\paragraph{\textbf{Baselines}.}
We compare \method against five baselines, on top of standard query and positive, the hard negatives are configured as: \textbf{Random} (15 randomly sampled corpus negatives), \textbf{Mined-only} (15 BM25-retrieved negatives), \textbf{Vanilla Generation} and \textbf{CoT Generation} (exclusively LLM-generated negatives using vanilla~\cite{li2024synegllmdrivensynthetichardnegatives} or CoT-guided generation), and \textbf{SyNeg} (na\"{i}ve mixing of 15 mined and 3 vanilla-generated negatives).
In comparison, our \textbf{\method} leverages on 3 CoT-generated negatives (\S\ref{sec:generation}), mixing with 15 mined ones with QEM strategy (\S\ref{sec:entropy}).
Detailed configurations are in Appendix~\ref{app:datasets}.

\subsection{Main Results}
\label{sec:main_results}

Table~\ref{tab:main_results} presents the main results across four retrieval benchmarks. Our key observations are as follows.

\paragraph{\textbf{Generated negatives without bridging the gap are destructive, not merely unhelpful.}}
Training exclusively on LLM-generated negatives leads to near-total collapse (average NDCG@10: 1.71 for Vanilla Generation, 11.45 for CoT Generation), confirming that the failure modes identified in Section~\ref{sec:analysis} have severe practical consequences.
Notably, CoT Generation substantially outperforms Vanilla Generation on mMARCO-zh (34.58 vs.\ 0.00) and HotpotQA (11.17 vs.\ 6.82), suggesting that structured reasoning partially mitigates discriminative-agnostic generation by producing negatives with more targeted semantic violations.
However, even CoT Generation remains catastrophic on NQ and TQA (NDCG@10 $< 0.1$), demonstrating that \emph{improving generation quality alone is insufficient}: when source-dependent shortcuts dominate the training signal, the model learns to discriminate by distributional artifacts regardless of the negatives' semantic content.

\paragraph{\textbf{Na\"{\i}ve mixing is unreliable without shortcut suppression}.}
SyNeg combines mined and vanilla-generated negatives without regularization, yielding inconsistent results. The gain is modest on mMARCO ($+0.49$ NDCG), but the degradation is substantial on TQA ($-3.59$) and NQ ($-2.82$).
This is directly predicted by our mutual information analysis (\S\ref{ssec:source_shortcuts}): when source identity is easily predictable from similarity scores, the model exploits this shortcut, and gradient drift corrupts the retrieval objective. The severity varies by how large the distributional gap is between generated and corpus text on each benchmark.
That 3 unregularized generated negatives can overwhelm 15 mined ones underscores how rapidly shortcut learning dominates optimization.

\paragraph{\textbf{\method validates the complementary design}.}
\method achieves the best NDCG@10 on all four benchmarks, with an average gain of $+1.90$ over Mined-only and $+3.50$ over SyNeg.
The improvement is most striking on TQA, where \method raises NDCG@10 from 56.75 to 59.02 ($+2.27$) while SyNeg \emph{degrades} performance to 53.16. This indicates a swing of $+5.86$ between regulated and unregulated incorporation of generated negatives.
Unlike SyNeg, which helps on some datasets and hurts on others, \method delivers consistent gains across all benchmarks, spanning Chinese web search (mMARCO-zh), multi-hop reasoning (HotpotQA), and factoid QA (NQ, TQA). This consistency confirms that both components are necessary and complementary: CoT-guided counterfactual perturbation ensures generated negatives carry genuine contrastive signal, while Query-View Entropy Maximization prevents the training process from bypassing that signal through source-identity shortcuts.

\subsection{Validation of Failure Mode Mitigation}
\label{sec:validation}

To validate that \method directly addresses the two failure modes diagnosed in Section~\ref{sec:analysis}---\emph{discriminative-agnostic generation} (Section~\ref{ssec:discriminative_agnostic}) and \emph{source-dependent shortcuts} (Section~\ref{ssec:source_shortcuts})---we evaluate both the generation-time embedding geometry of negatives and their training-time gradient dynamics under mixed negatives.

\paragraph{Mitigating discriminative-agnostic generation.}
We repeat the embedding space diagnostics using a frozen Qwen3-Embedding-0.6B encoder. Under vanilla generation, generated negatives exhibit a strong source fingerprint: 24\% fall into HDBSCAN \emph{pure clusters} containing only generated negatives. CoT-guided counterfactual perturbation largely eliminates this artifact, reducing the pure-cluster ratio to 3\% and producing negatives that intermix with queries, positives, and mined negatives (Appendix~\ref{app:embedding_comparison}). Table~\ref{tab:quality_diversity} further shows that CoT generation narrows the query-similarity gap to mined negatives (0.516 vs 0.529 for mined; vanilla: 0.631), increases similarity-range overlap with mined (0.890 vs 0.660), improves intra-negative diversity (0.512 vs 0.321), and reduces template-pattern repetition (0.2\% vs 1.1\%). Together, these results confirm that CoT generation replaces generic/topic-drifted negatives with semantically grounded, corpus-like hard negatives that better match the discriminative requirements.

\paragraph{Mitigating source-dependent shortcuts.}
We then test whether generated negatives remain effective when mixed with mined negatives, focusing on the shortcut-induced collapse and gradient drift in Section~\ref{ssec:source_shortcuts}. We track each negative's softmax weight in the InfoNCE denominator (i.e., per-negative gradient contribution) and gradient statistics across checkpoints for SyNeg (15 mined + 3 vanilla) and \method (15 mined + 3 CoT + QEM). At checkpoint-10, SyNeg’s generated negatives contribute almost no gradient (generated/mined per-negative weight ratio 0.008) and are substantially less similar to the query than mined negatives (0.480 vs 0.564), consistent with early shortcut-driven collapse. In contrast, \method maintains meaningful early learning signal from generated negatives (weight ratio 0.80; similarity 0.603 vs 0.583 for mined). As training progresses, while both methods eventually down-weight generated negatives, \method dampens residual interference: at checkpoint-100 the generated/mined gradient-norm ratio is 0.35 for \method versus 0.58 for SyNeg, and SyNeg’s mined-generated gradient alignment drops to 0.24. This shows that \method not only prevents early collapse but also reduces shortcut-driven gradient drift once generated negatives become easy.

\begin{table}[t]
    \centering
    \caption{Embedding space quality and diversity comparison between vanilla and CoT-generated negatives.}
    \vspace{-1em}
    \label{tab:quality_diversity}
    \begin{tabular}{lcc}
        \toprule
        Metric & Vanilla & CoT \\
        \midrule
Pure cluster ratio (\%) & 24 & 3 \\
Mean query similarity & 0.631 & 0.516 \\
Intra-negative diversity & 0.321 & 0.512 \\
Similarity range overlap with mined & 0.660 & 0.890 \\
Template pattern ratio (\%) & 1.1 & 0.2 \\

        \bottomrule
    \end{tabular}
\end{table}

\subsection{Ablation Study}
\label{sec:ablation}

We conduct ablation experiments on TQA to isolate the contribution of each component.

\begin{table}[t]
    \centering
    \caption{Ablation study on TQA. Both CoT-guided generation and entropy regularization contribute to performance.}
    \vspace{-1em}
    \label{tab:ablation}
    \begin{tabular}{lcc}
        \toprule
        Configuration & NDCG@10 & Recall@10 \\
        \midrule
        \method & \textbf{59.02} & \textbf{67.11} \\
        \midrule
        \quad w/o entropy loss & 57.61 & 65.00 \\
        \quad w/o CoT generation & 57.11 & 65.45 \\
        \bottomrule
    \end{tabular}
\end{table}

Removing entropy regularization while retaining CoT-generated negatives (``w/o entropy loss'') reduces NDCG@10 from 59.02 to 57.61, a gain of only $+0.86$ over Mined-only (56.75). Removing CoT generation while retaining entropy regularization with vanilla negatives (``w/o CoT generation'') yields 57.11. Neither component alone matches the full system, confirming that improved generation and explicit regularization provide complementary benefits: CoT generation reduces the distributional gap (Section~\ref{sec:validation}), while entropy regularization suppresses exploitation of any residual divergence.

\subsection{Scaling of Generated Negatives} 
\label{sec:scaling}

We examine how the quantity of CoT-generated negatives affects retrieval performance on TQA along two dimensions: the proportion of queries receiving generated augmentation, and the number of generated negatives per query. Sensitivity analysis of QEM hyperparameters ($\tau_{\text{ent}}$, $\lambda$) and chain position selection strategies is provided in Appendix~\ref{app:sensitivity}.

\begin{figure}[t]
    \centering
    \includegraphics[width=\columnwidth]{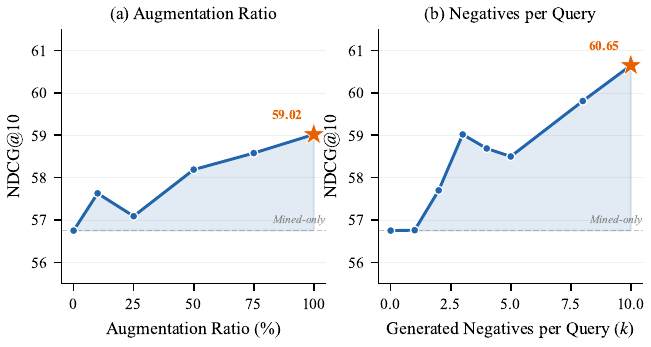}
    \vspace{-2.5em}
    \caption{Scaling behavior of CoT-generated negatives on TQA. (a)~Performance improves monotonically with the proportion of queries augmented. (b)~Adding more generated negatives per query ($k$) overall trend improves performance. Dashed lines: Mined-only baseline; stars: best configuration.}
    \label{fig:scaling}
\end{figure}

\paragraph{Augmentation ratio.}
Figure~\ref{fig:scaling}(a) shows results when varying the proportion of queries augmented with generated negatives from 0\% to 100\%, with $\tau_{\text{ent}} = 0.1$ and $\lambda = 0.1$ fixed. Performance improves monotonically with the ratio, reaching the best result at 100\%. Even augmenting only 10\% of queries yields a meaningful $+0.88$ NDCG@10 improvement over the Mined-only baseline, demonstrating practical cost-efficiency: in resource-constrained settings, selectively generating negatives for a fraction of queries can still yield substantial gains.

\paragraph{Number of generated negatives.}
Figure~\ref{fig:scaling}(b) reports performance when varying the number of CoT-generated negatives per query from 0 (Mined-only) to 10 while keeping 15 mined negatives fixed. The overall trend is clear: more generated negatives yield better performance, with $k{=}10$ achieving 60.65 NDCG@10 ($+3.90$ over Mined-only, $+1.63$ over the $k{=}3$ default). The steep improvement from $k{=}1$ to $k{=}3$ suggests that a minimum diversity of perturbation types is necessary for effective training. Performance at $k{=}4$ and $k{=}5$ dips slightly, likely due to negatives from less critical chain nodes; however, at $k{=}8$ and $k{=}10$, the benefits of additional coverage outweigh this effect. These results indicate that the CoT pipeline scales favorably: investing in more generated negatives per query translates to continued retrieval gains.

\section{Related Work}
\label{sec:related}

\paragraph{Text Embeddings and Contrastive Learning.}
Modern dense retrievers~\cite{karpukhin-etal-2020-dense, qwen3embedding, lee2025geminiembeddinggeneralizableembeddings} are typically initialized from pre-trained LLMs and fine-tuned with contrastive objectives such as InfoNCE~\cite{robinson2021contrastive}, where negative quality is critical. Simple in-batch random negatives~\cite{karpukhin-etal-2020-dense, xiong2021approximate}, while computationally efficient, often prove insufficiently challenging, motivating hard negative mining via BM25~\cite{robertson1994some}, cross-encoder reranking~\cite{moreira2024nvretrieverimprovingtextembedding}, or dense retrieval with frozen or iteratively updated encoders. Recent work has introduced principled sampling strategies: SimANS~\cite{zhou-etal-2022-simans} targets ambiguous negatives near the positive boundary, and TriSampler~\cite{TriSampler} formalizes geometric constraints among queries, positives, and negatives. Despite these advances, all mining-based approaches are fundamentally bounded by corpus availability, limiting coverage of underrepresented boundary regions.

\paragraph{LLM-based Data Synthesis and Its Risks.}
LLMs have been widely adopted for synthesizing training data across NLP tasks~\cite{long-etal-2024-llms}, including generating queries, passages, or query-document pairs for embedding training~\cite{wang2024improving}. SyNeg~\cite{li2024synegllmdrivensynthetichardnegatives} specifically targets hard negative generation conditioned on queries and positives. However, training on synthetic data carries systematic risks: model collapse~\cite{shumailov2024ai, gan2025towards} degrades diversity through iterative self-training, and distributional shifts from synthetic augmentation are well documented~\cite{GenDataAgent, jung2025prismatic, nguyen2025provably,ToEdit,ren2025fewshotllmsyntheticdata}. Our work identifies a distinct failure mode in the contrastive learning setting: even single-round LLM-generated negatives introduce distributional artifacts that enable shortcut learning and cause gradient drift away from the relevance boundary. To our knowledge, this is the first study to systematically diagnose both the generation-level and optimization-level consequences of synthetic negatives for contrastive retrieval training.

\section{Conclusion}
\label{sec:conclusion}

We identify and formalize a \emph{generative-discriminative gap} explaining why na\"{i}vely incorporating LLM-generated hard negatives into contrastive learning degrades retrieval performance. We trace this gap to two compounding failure modes: \emph{discriminative-agnostic generation} and \emph{source-dependent shortcuts}.
To close this gap, we proposed \method, intervening at both pipeline stages. \emph{CoT-guided counterfactual perturbation} decomposes query-document relevance into explicit information requirements and surgically violates individual ones, shifting from accidental to intentional hardness. \emph{Query-view entropy maximization} disperses generated negatives across the similarity spectrum, suppressing shortcut exploitation by minimizing mutual information between source identity and similarity scores. Experiments across multiple benchmarks confirm that these components are complementary and that \method consistently outperforms mining-only and na\"{i}ve generation baselines.
More broadly, our findings expose a general risk in augmenting discriminative objectives with generative synthetic data. 
As LLM-generated content grows prevalent in training corpora, principled alignment of synthetic data with discriminative requirements will become increasingly important.

\begin{acks}
This research is supported by National Natural Science Foundation of China (Grant No.62276154);
the Natural Science Foundation of Guangdong Province (Grant No.2024TQ08X729);
Basic Research Fund of Shenzhen City (Grant No.JCYJ20240813112009013 and GJHZ20240218113603006);
The Major Key Project of PCL for Experiments and Applications (Grant No.PCL2024A08).
\end{acks}

\bibliographystyle{ACM-Reference-Format}
\bibliography{ref}

@String{Computing = "Computing" }

@String{Computer = "{IEEE} Computer" }

@String{Springer = "Springer-Verlag" }

@article{qwen3embedding,
  title={Qwen3 Embedding: Advancing Text Embedding and Reranking Through Foundation Models},
  author={Zhang, Yanzhao and Li, Mingxin and Long, Dingkun and Zhang, Xin and Lin, Huan and Yang, Baosong and Xie, Pengjun and Yang, An and Liu, Dayiheng and Lin, Junyang and Huang, Fei and Zhou, Jingren},
  journal={arXiv preprint arXiv:2506.05176},
  year={2025}
}

@inproceedings{karpukhin-etal-2020-dense,
    title = "Dense Passage Retrieval for Open-Domain Question Answering",
    author = "Karpukhin, Vladimir  and
      Oguz, Barlas  and
      Min, Sewon  and
      Lewis, Patrick  and
      Wu, Ledell  and
      Edunov, Sergey  and
      Chen, Danqi  and
      Yih, Wen-tau",
    editor = "Webber, Bonnie  and
      Cohn, Trevor  and
      He, Yulan  and
      Liu, Yang",
    booktitle = "Proceedings of the 2020 Conference on Empirical Methods in Natural Language Processing (EMNLP)",
    month = nov,
    year = "2020",
    address = "Online",
    publisher = "Association for Computational Linguistics",
    url = "https://aclanthology.org/2020.emnlp-main.550/",
    doi = "10.18653/v1/2020.emnlp-main.550",
    pages = "6769--6781"
}

@inproceedings{moreira2024nvretrieverimprovingtextembedding,
author = {Moreira, Gabriel de Souza P. and Osmulski, Radek and Xu, Mengyao and Ak, Ronay and Schifferer, Benedikt and Oldridge, Even},
title = {Improving Text Embedding Models with Positive-aware Hard-negative Mining},
year = {2025},
isbn = {9798400720406},
publisher = {Association for Computing Machinery},
address = {New York, NY, USA},
url = {https://doi.org/10.1145/3746252.3761254},
doi = {10.1145/3746252.3761254},
booktitle = {Proceedings of the 34th ACM International Conference on Information and Knowledge Management},
pages = {2169–2178},
numpages = {10},
location = {Seoul, Republic of Korea},
series = {CIKM '25}
}

@inproceedings{wei2023chainofthoughtpromptingelicitsreasoning,
author = {Wei, Jason and Wang, Xuezhi and Schuurmans, Dale and Bosma, Maarten and Ichter, Brian and Xia, Fei and Chi, Ed H. and Le, Quoc V. and Zhou, Denny},
title = {Chain-of-thought prompting elicits reasoning in large language models},
year = {2022},
isbn = {9781713871088},
publisher = {Curran Associates Inc.},
address = {Red Hook, NY, USA},
booktitle = {Proceedings of the 36th International Conference on Neural Information Processing Systems},
articleno = {1800},
numpages = {14},
location = {New Orleans, LA, USA},
series = {NIPS '22}
}

@misc{li2024synegllmdrivensynthetichardnegatives,
      title={SyNeg: LLM-Driven Synthetic Hard-Negatives for Dense Retrieval}, 
      author={Xiaopeng Li and Xiangyang Li and Hao Zhang and Zhaocheng Du and Pengyue Jia and Yichao Wang and Xiangyu Zhao and Huifeng Guo and Ruiming Tang},
      year={2024},
      eprint={2412.17250},
      archivePrefix={arXiv},
      primaryClass={cs.IR},
      url={https://arxiv.org/abs/2412.17250}, 
}

@inproceedings{jia-etal-2025-bridging,
    title = "Bridging Relevance and Reasoning: Rationale Distillation in Retrieval-Augmented Generation",
    author = "Jia, Pengyue  and
      Xu, Derong  and
      Li, Xiaopeng  and
      Du, Zhaocheng  and
      Li, Xiangyang  and
      Wang, Yichao  and
      Wang, Yuhao  and
      Liu, Qidong  and
      Wang, Maolin  and
      Guo, Huifeng  and
      Tang, Ruiming  and
      Zhao, Xiangyu",
    editor = "Che, Wanxiang  and
      Nabende, Joyce  and
      Shutova, Ekaterina  and
      Pilehvar, Mohammad Taher",
    booktitle = "Findings of the Association for Computational Linguistics: ACL 2025",
    month = jul,
    year = "2025",
    address = "Vienna, Austria",
    publisher = "Association for Computational Linguistics",
    url = "https://aclanthology.org/2025.findings-acl.220/",
    doi = "10.18653/v1/2025.findings-acl.220",
    pages = "4242--4256",
    ISBN = "979-8-89176-256-5"
}

@inproceedings{
robinson2021contrastive,
title={Contrastive Learning with Hard Negative Samples},
author={Joshua David Robinson and Ching-Yao Chuang and Suvrit Sra and Stefanie Jegelka},
booktitle={International Conference on Learning Representations},
year={2021},
url={https://openreview.net/forum?id=CR1XOQ0UTh-}
}

@inproceedings{swayamdipta-etal-2020-dataset,
    title = "Dataset Cartography: Mapping and Diagnosing Datasets with Training Dynamics",
    author = "Swayamdipta, Swabha  and
      Schwartz, Roy  and
      Lourie, Nicholas  and
      Wang, Yizhong  and
      Hajishirzi, Hannaneh  and
      Smith, Noah A.  and
      Choi, Yejin",
    editor = "Webber, Bonnie  and
      Cohn, Trevor  and
      He, Yulan  and
      Liu, Yang",
    booktitle = "Proceedings of the 2020 Conference on Empirical Methods in Natural Language Processing (EMNLP)",
    month = nov,
    year = "2020",
    address = "Online",
    publisher = "Association for Computational Linguistics",
    url = "https://aclanthology.org/2020.emnlp-main.746/",
    doi = "10.18653/v1/2020.emnlp-main.746",
    pages = "9275--9293"
}

@InProceedings{pmlr-v162-zhong22a,
  title = 	 {Describing Differences between Text Distributions with Natural Language},
  author =       {Zhong, Ruiqi and Snell, Charlie and Klein, Dan and Steinhardt, Jacob},
  booktitle = 	 {Proceedings of the 39th International Conference on Machine Learning},
  pages = 	 {27099--27116},
  year = 	 {2022},
  editor = 	 {Chaudhuri, Kamalika and Jegelka, Stefanie and Song, Le and Szepesvari, Csaba and Niu, Gang and Sabato, Sivan},
  volume = 	 {162},
  series = 	 {Proceedings of Machine Learning Research},
  month = 	 {17--23 Jul},
  publisher =    {PMLR},
  url = 	 {https://proceedings.mlr.press/v162/zhong22a.html}
}

@inproceedings{singh-etal-2023-explaining,
    title = "Explaining Data Patterns in Natural Language with Language Models",
    author = "Singh, Chandan  and
      Morris, John X.  and
      Aneja, Jyoti  and
      Rush, Alexander  and
      Gao, Jianfeng",
    editor = "Belinkov, Yonatan  and
      Hao, Sophie  and
      Jumelet, Jaap  and
      Kim, Najoung  and
      McCarthy, Arya  and
      Mohebbi, Hosein",
    booktitle = "Proceedings of the 6th BlackboxNLP Workshop: Analyzing and Interpreting Neural Networks for NLP",
    month = dec,
    year = "2023",
    address = "Singapore",
    publisher = "Association for Computational Linguistics",
    url = "https://aclanthology.org/2023.blackboxnlp-1.3/",
    doi = "10.18653/v1/2023.blackboxnlp-1.3",
    pages = "31--55"
}

@inproceedings{GenDataAgent,
 author = {Li, Zhiteng and Chen, Lele and Andrews, Jerone and Ba, Yunhao and Zhang, Yulun and Xiang, Alice},
 booktitle = {International Conference on Learning Representations},
 editor = {Y. Yue and A. Garg and N. Peng and F. Sha and R. Yu},
 pages = {48578--48598},
 title = {GenDataAgent: On-the-fly Dataset Augmentation with Synthetic Data},
 url = {https://proceedings.iclr.cc/paper_files/paper/2025/file/79081c95482707d2db390542614e29cd-Paper-Conference.pdf},
 volume = {2025},
 year = {2025}
}

@inproceedings{
ToEdit,
title={How to Synthesize Text Data without Model Collapse?},
author={Xuekai Zhu and Daixuan Cheng and Hengli Li and Kaiyan Zhang and Ermo Hua and Xingtai Lv and Ning Ding and Zhouhan Lin and Zilong Zheng and Bowen Zhou},
booktitle={Forty-second International Conference on Machine Learning},
year={2025},
url={https://openreview.net/forum?id=ihUi76a4u7}
}

@inproceedings{wang2024improving,
  title={Improving text embeddings with large language models},
  author={Wang, Liang and Yang, Nan and Huang, Xiaolong and Yang, Linjun and Majumder, Rangan and Wei, Furu},
  booktitle={Proceedings of the 62nd Annual Meeting of the Association for Computational Linguistics (Volume 1: Long Papers)},
  pages={11897--11916},
  year={2024}
}

@inproceedings{long-etal-2024-llms,
    title = "On {LLM}s-Driven Synthetic Data Generation, Curation, and Evaluation: A Survey",
    author = "Long, Lin  and
      Wang, Rui  and
      Xiao, Ruixuan  and
      Zhao, Junbo  and
      Ding, Xiao  and
      Chen, Gang  and
      Wang, Haobo",
    editor = "Ku, Lun-Wei  and
      Martins, Andre  and
      Srikumar, Vivek",
    booktitle = "Findings of the Association for Computational Linguistics: ACL 2024",
    month = aug,
    year = "2024",
    address = "Bangkok, Thailand",
    publisher = "Association for Computational Linguistics",
    url = "https://aclanthology.org/2024.findings-acl.658/",
    doi = "10.18653/v1/2024.findings-acl.658",
    pages = "11065--11082"
}

@inproceedings{
jung2025prismatic,
title={Prismatic Synthesis: Gradient-based Data Diversification Boosts Generalization in {LLM} Reasoning},
author={Jaehun Jung and Seungju Han and Ximing Lu and Skyler Hallinan and David Acuna and Shrimai Prabhumoye and Mostofa Patwary and Mohammad Shoeybi and Bryan Catanzaro and Yejin Choi},
booktitle={The Thirty-ninth Annual Conference on Neural Information Processing Systems},
year={2025},
url={https://openreview.net/forum?id=R0dC7Xzwbk}
}

@inproceedings{
nguyen2025provably,
title={Provably Improving Generalization of Few-shot models with Synthetic Data},
author={Lan-Cuong Nguyen and Quan Nguyen-Tri and Bang Tran Khanh and Dung D. Le and Long Tran-Thanh and Khoat Than},
booktitle={Forty-second International Conference on Machine Learning},
year={2025},
url={https://openreview.net/forum?id=L6U7nYc4ah}
}

@inproceedings{zhou-etal-2022-simans,
    title = "{S}im{ANS}: Simple Ambiguous Negatives Sampling for Dense Text Retrieval",
    author = "Zhou, Kun  and
      Gong, Yeyun  and
      Liu, Xiao  and
      Zhao, Wayne Xin  and
      Shen, Yelong  and
      Dong, Anlei  and
      Lu, Jingwen  and
      Majumder, Rangan  and
      Wen, Ji-rong  and
      Duan, Nan",
    editor = "Li, Yunyao  and
      Lazaridou, Angeliki",
    booktitle = "Proceedings of the 2022 Conference on Empirical Methods in Natural Language Processing: Industry Track",
    month = dec,
    year = "2022",
    address = "Abu Dhabi, UAE",
    publisher = "Association for Computational Linguistics",
    url = "https://aclanthology.org/2022.emnlp-industry.56/",
    doi = "10.18653/v1/2022.emnlp-industry.56",
    pages = "548--559"
}

@inproceedings{TriSampler,
author = {Yang, Zhen and Shao, Zhou and Dong, Yuxiao and Tang, Jie},
title = {TriSampler: a better negative sampling principle for dense retrieval},
year = {2024},
isbn = {978-1-57735-887-9},
publisher = {AAAI Press},
url = {https://doi.org/10.1609/aaai.v38i8.28779},
doi = {10.1609/aaai.v38i8.28779},
booktitle = {Proceedings of the Thirty-Eighth AAAI Conference on Artificial Intelligence and Thirty-Sixth Conference on Innovative Applications of Artificial Intelligence and Fourteenth Symposium on Educational Advances in Artificial Intelligence},
articleno = {1031},
numpages = {9},
series = {AAAI'24/IAAI'24/EAAI'24}
}

@inproceedings{marco,
  author       = {Tri Nguyen and
                  Mir Rosenberg and
                  Xia Song and
                  Jianfeng Gao and
                  Saurabh Tiwary and
                  Rangan Majumder and
                  Li Deng},
  editor       = {Tarek Richard Besold and
                  Antoine Bordes and
                  Artur S. d'Avila Garcez and
                  Greg Wayne},
  title        = {{MS} {MARCO:} {A} Human Generated MAchine Reading COmprehension Dataset},
  booktitle    = {Proceedings of the Workshop on Cognitive Computation: Integrating
                  neural and symbolic approaches 2016 co-located with the 30th Annual
                  Conference on Neural Information Processing Systems {(NIPS} 2016),
                  Barcelona, Spain, December 9, 2016},
  series       = {{CEUR} Workshop Proceedings},
  volume       = {1773},
  publisher    = {CEUR-WS.org},
  year         = {2016},
  url          = {https://ceur-ws.org/Vol-1773/CoCoNIPS\_2016\_paper9.pdf},
  bibsource    = {dblp computer science bibliography, https://dblp.org}
}

@article{kwiatkowski-etal-2019-natural,
    title = "Natural Questions: A Benchmark for Question Answering Research",
    author = "Kwiatkowski, Tom  and
      Palomaki, Jennimaria  and
      Redfield, Olivia  and
      Collins, Michael  and
      Parikh, Ankur  and
      Alberti, Chris  and
      Epstein, Danielle  and
      Polosukhin, Illia  and
      Devlin, Jacob  and
      Lee, Kenton  and
      Toutanova, Kristina  and
      Jones, Llion  and
      Kelcey, Matthew  and
      Chang, Ming-Wei  and
      Dai, Andrew M.  and
      Uszkoreit, Jakob  and
      Le, Quoc  and
      Petrov, Slav",
    editor = "Lee, Lillian  and
      Johnson, Mark  and
      Roark, Brian  and
      Nenkova, Ani",
    journal = "Transactions of the Association for Computational Linguistics",
    volume = "7",
    year = "2019",
    address = "Cambridge, MA",
    publisher = "MIT Press",
    url = "https://aclanthology.org/Q19-1026/",
    doi = "10.1162/tacl_a_00276",
    pages = "452--466"
}

@inproceedings{joshi-etal-2017-triviaqa,
    title = "{T}rivia{QA}: A Large Scale Distantly Supervised Challenge Dataset for Reading Comprehension",
    author = "Joshi, Mandar  and
      Choi, Eunsol  and
      Weld, Daniel  and
      Zettlemoyer, Luke",
    editor = "Barzilay, Regina  and
      Kan, Min-Yen",
    booktitle = "Proceedings of the 55th Annual Meeting of the Association for Computational Linguistics (Volume 1: Long Papers)",
    month = jul,
    year = "2017",
    address = "Vancouver, Canada",
    publisher = "Association for Computational Linguistics",
    url = "https://aclanthology.org/P17-1147/",
    doi = "10.18653/v1/P17-1147",
    pages = "1601--1611"
}

@inproceedings{yang-etal-2018-hotpotqa,
    title = "{H}otpot{QA}: A Dataset for Diverse, Explainable Multi-hop Question Answering",
    author = "Yang, Zhilin  and
      Qi, Peng  and
      Zhang, Saizheng  and
      Bengio, Yoshua  and
      Cohen, William  and
      Salakhutdinov, Ruslan  and
      Manning, Christopher D.",
    editor = "Riloff, Ellen  and
      Chiang, David  and
      Hockenmaier, Julia  and
      Tsujii, Jun{'}ichi",
    booktitle = "Proceedings of the 2018 Conference on Empirical Methods in Natural Language Processing",
    month = oct # "-" # nov,
    year = "2018",
    address = "Brussels, Belgium",
    publisher = "Association for Computational Linguistics",
    url = "https://aclanthology.org/D18-1259/",
    doi = "10.18653/v1/D18-1259",
    pages = "2369--2380"
}

@inproceedings{chen-etal-2024-m3,
    title = "{M}3-Embedding: Multi-Linguality, Multi-Functionality, Multi-Granularity Text Embeddings Through Self-Knowledge Distillation",
    author = "Chen, Jianlyu  and
      Xiao, Shitao  and
      Zhang, Peitian  and
      Luo, Kun  and
      Lian, Defu  and
      Liu, Zheng",
    editor = "Ku, Lun-Wei  and
      Martins, Andre  and
      Srikumar, Vivek",
    booktitle = "Findings of the Association for Computational Linguistics: ACL 2024",
    month = aug,
    year = "2024",
    address = "Bangkok, Thailand",
    publisher = "Association for Computational Linguistics",
    url = "https://aclanthology.org/2024.findings-acl.137/",
    doi = "10.18653/v1/2024.findings-acl.137",
    pages = "2318--2335"
}

@misc{lee2025geminiembeddinggeneralizableembeddings,
      title={Gemini Embedding: Generalizable Embeddings from Gemini}, 
      author={Jinhyuk Lee and Feiyang Chen and Sahil Dua and Daniel Cer and Madhuri Shanbhogue and Iftekhar Naim and Gustavo Hernández Ábrego and Zhe Li and Kaifeng Chen and Henrique Schechter Vera and Xiaoqi Ren and Shanfeng Zhang and Daniel Salz and Michael Boratko and Jay Han and Blair Chen and Shuo Huang and Vikram Rao and Paul Suganthan and Feng Han and Andreas Doumanoglou and Nithi Gupta and Fedor Moiseev and Cathy Yip and Aashi Jain and Simon Baumgartner and Shahrokh Shahi and Frank Palma Gomez and Sandeep Mariserla and Min Choi and Parashar Shah and Sonam Goenka and Ke Chen and Ye Xia and Koert Chen and Sai Meher Karthik Duddu and Yichang Chen and Trevor Walker and Wenlei Zhou and Rakesh Ghiya and Zach Gleicher and Karan Gill and Zhe Dong and Mojtaba Seyedhosseini and Yunhsuan Sung and Raphael Hoffmann and Tom Duerig},
      year={2025},
      eprint={2503.07891},
      archivePrefix={arXiv},
      primaryClass={cs.CL},
      url={https://arxiv.org/abs/2503.07891}, 
}

@inproceedings{
xiong2021approximate,
title={Approximate Nearest Neighbor Negative Contrastive Learning for Dense Text Retrieval},
author={Lee Xiong and Chenyan Xiong and Ye Li and Kwok-Fung Tang and Jialin Liu and Paul N. Bennett and Junaid Ahmed and Arnold Overwijk},
booktitle={International Conference on Learning Representations},
year={2021},
url={https://openreview.net/forum?id=zeFrfgyZln}
}

@inproceedings{robertson1994some,
  title={Some simple effective approximations to the 2-poisson model for probabilistic weighted retrieval},
  author={Robertson, Stephen E and Walker, Steve},
  booktitle={SIGIR’94: Proceedings of the Seventeenth Annual International ACM-SIGIR Conference on Research and Development in Information Retrieval, organised by Dublin City University},
  pages={232--241},
  year={1994},
  organization={Springer}
}

@inproceedings{campello2013density,
  title={Density-based clustering based on hierarchical density estimates},
  author={Campello, Ricardo JGB and Moulavi, Davoud and Sander, J{\"o}rg},
  booktitle={Pacific-Asia conference on knowledge discovery and data mining},
  pages={160--172},
  year={2013},
  organization={Springer}
}

@inproceedings{swift,
author = {Zhao, Yuze and Huang, Jintao and Hu, Jinghan and Wang, Xingjun and Mao, Yunlin and Zhang, Daoze and Jiang, Zeyinzi and Wu, Zhikai and Ai, Baole and Wang, Ang and Zhou, Wenmeng and Chen, Yingda},
title = {SWIFT: a scalable lightweight infrastructure for fine-tuning},
year = {2025},
isbn = {978-1-57735-897-8},
publisher = {AAAI Press},
url = {https://doi.org/10.1609/aaai.v39i28.35383},
doi = {10.1609/aaai.v39i28.35383},
booktitle = {Proceedings of the Thirty-Ninth AAAI Conference on Artificial Intelligence and Thirty-Seventh Conference on Innovative Applications of Artificial Intelligence and Fifteenth Symposium on Educational Advances in Artificial Intelligence},
articleno = {3485},
numpages = {3},
series = {AAAI'25/IAAI'25/EAAI'25}
}

@inproceedings{yates-etal-2021-pretrained,
    title = "Pretrained Transformers for Text Ranking: {BERT} and Beyond",
    author = "Yates, Andrew  and
      Nogueira, Rodrigo  and
      Lin, Jimmy",
    editor = "Kondrak, Greg  and
      Bontcheva, Kalina  and
      Gillick, Dan",
    booktitle = "Proceedings of the 2021 Conference of the North American Chapter of the Association for Computational Linguistics: Human Language Technologies: Tutorials",
    month = jun,
    year = "2021",
    address = "Online",
    publisher = "Association for Computational Linguistics",
    url = "https://aclanthology.org/2021.naacl-tutorials.1/",
    doi = "10.18653/v1/2021.naacl-tutorials.1",
    pages = "1--4"
}

@misc{oord2019representationlearningcontrastivepredictive,
      title={Representation Learning with Contrastive Predictive Coding}, 
      author={Aaron van den Oord and Yazhe Li and Oriol Vinyals},
      year={2019},
      eprint={1807.03748},
      archivePrefix={arXiv},
      primaryClass={cs.LG},
      url={https://arxiv.org/abs/1807.03748}, 
}

@inproceedings{10.5555/3524938.3525087,
author = {Chen, Ting and Kornblith, Simon and Norouzi, Mohammad and Hinton, Geoffrey},
title = {A simple framework for contrastive learning of visual representations},
year = {2020},
publisher = {JMLR.org},
booktitle = {Proceedings of the 37th International Conference on Machine Learning},
articleno = {149},
numpages = {11},
series = {ICML'20}
}

@inproceedings{
    thakur2021beir,
    title={{BEIR}: A Heterogeneous Benchmark for Zero-shot Evaluation of Information Retrieval Models},
    author={Nandan Thakur and Nils Reimers and Andreas R{\"u}ckl{\'e} and Abhishek Srivastava and Iryna Gurevych},
    booktitle={Thirty-fifth Conference on Neural Information Processing Systems Datasets and Benchmarks Track (Round 2)},
    year={2021},
    url={https://openreview.net/forum?id=wCu6T5xFjeJ}
}

@inproceedings{qu-etal-2021-rocketqa,
    title = "{R}ocket{QA}: An Optimized Training Approach to Dense Passage Retrieval for Open-Domain Question Answering",
    author = "Qu, Yingqi  and
      Ding, Yuchen  and
      Liu, Jing  and
      Liu, Kai  and
      Ren, Ruiyang  and
      Zhao, Wayne Xin  and
      Dong, Daxiang  and
      Wu, Hua  and
      Wang, Haifeng",
    editor = "Toutanova, Kristina  and
      Rumshisky, Anna  and
      Zettlemoyer, Luke  and
      Hakkani-Tur, Dilek  and
      Beltagy, Iz  and
      Bethard, Steven  and
      Cotterell, Ryan  and
      Chakraborty, Tanmoy  and
      Zhou, Yichao",
    booktitle = "Proceedings of the 2021 Conference of the North American Chapter of the Association for Computational Linguistics: Human Language Technologies",
    month = jun,
    year = "2021",
    address = "Online",
    publisher = "Association for Computational Linguistics",
    url = "https://aclanthology.org/2021.naacl-main.466/",
    doi = "10.18653/v1/2021.naacl-main.466",
    pages = "5835--5847"
}

@inproceedings{10.1145/3404835.3462880,
author = {Zhan, Jingtao and Mao, Jiaxin and Liu, Yiqun and Guo, Jiafeng and Zhang, Min and Ma, Shaoping},
title = {Optimizing Dense Retrieval Model Training with Hard Negatives},
year = {2021},
isbn = {9781450380379},
publisher = {Association for Computing Machinery},
address = {New York, NY, USA},
url = {https://doi.org/10.1145/3404835.3462880},
doi = {10.1145/3404835.3462880},
booktitle = {Proceedings of the 44th International ACM SIGIR Conference on Research and Development in Information Retrieval},
pages = {1503–1512},
numpages = {10},
location = {Virtual Event, Canada},
series = {SIGIR '21}
}

@inproceedings{inpars,
  author = {Bonifacio, Luiz and Abonizio, Hugo and Fadaee, Marzieh and Nogueira, Rodrigo},
  title = {{InPars}: Unsupervised Dataset Generation for Information Retrieval},
  year = {2022},
  isbn = {9781450387323},
  publisher = {Association for Computing Machinery},
  address = {New York, NY, USA},
  url = {https://doi.org/10.1145/3477495.3531863},
  doi = {10.1145/3477495.3531863},
  booktitle = {Proceedings of the 45th International ACM SIGIR Conference on Research and Development in Information Retrieval},
  pages = {2387–2392},
  numpages = {6},
  location = {Madrid, Spain},
  series = {SIGIR '22}
}

@article{Bonifacio2021MMarco,
    title={{mMARCO}: A Multilingual Version of {MS MARCO} Passage Ranking Dataset},
    author={Luiz Henrique Bonifacio and Israel Campiotti and Roberto Lotufo and Rodrigo Nogueira},
    year={2021},
    journal={arXiv:2108.13897}
}

@misc{dong2025mmdocrag,
      title={Benchmarking Retrieval-Augmented Multimodal Generation for Document Question Answering}, 
      author={Kuicai Dong and Yujing Chang and Shijie Huang and Yasheng Wang and Ruiming Tang and Yong Liu},
      year={2025},
      eprint={2505.16470},
      archivePrefix={arXiv},
      primaryClass={cs.IR},
      url={https://arxiv.org/abs/2505.16470}, 
}

@inproceedings{dong2025mmdocir,
    title = "{MMD}oc{IR}: Benchmarking Multimodal Retrieval for Long Documents",
    author = "Dong, Kuicai  and
      Chang, Yujing  and
      Goh Xin Deik, Derrick  and
      Li, Dexun  and
      Tang, Ruiming  and
      Liu, Yong",
    editor = "Christodoulopoulos, Christos  and
      Chakraborty, Tanmoy  and
      Rose, Carolyn  and
      Peng, Violet",
    booktitle = "Proceedings of the 2025 Conference on Empirical Methods in Natural Language Processing",
    month = nov,
    year = "2025",
    address = "Suzhou, China",
    publisher = "Association for Computational Linguistics",
    url = "https://aclanthology.org/2025.emnlp-main.1576/",
    doi = "10.18653/v1/2025.emnlp-main.1576",
    pages = "30971--31005",
    ISBN = "979-8-89176-332-6",
}

@inproceedings{tang-etal-2025-perception,
    title = "Perception Compressor: A Training-Free Prompt Compression Framework in Long Context Scenarios",
    author = "Tang, Jiwei  and
      Xu, Jin  and
      Lu, Tingwei  and
      Zhang, Zhicheng  and
      Zhao, Yiming  and
      Hai, Lin  and
      Zheng, Hai-Tao",
    editor = "Chiruzzo, Luis  and
      Ritter, Alan  and
      Wang, Lu",
    booktitle = "Findings of the Association for Computational Linguistics: NAACL 2025",
    month = apr,
    year = "2025",
    address = "Albuquerque, New Mexico",
    publisher = "Association for Computational Linguistics",
    url = "https://aclanthology.org/2025.findings-naacl.229/",
    doi = "10.18653/v1/2025.findings-naacl.229",
    pages = "4093--4108",
    ISBN = "979-8-89176-195-7"
}

@misc{tang2026gmsaenhancingcontextcompression,
      title={GMSA: Enhancing Context Compression via Group Merging and Layer Semantic Alignment}, 
      author={Jiwei Tang and Zhicheng Zhang and Shunlong Wu and Jingheng Ye and Lichen Bai and Zitai Wang and Tingwei Lu and Lin Hai and Yiming Zhao and Hai-Tao Zheng and Hong-Gee Kim},
      year={2026},
      eprint={2505.12215},
      archivePrefix={arXiv},
      primaryClass={cs.CL},
      url={https://arxiv.org/abs/2505.12215}, 
}

@inproceedings{
tang2026comi,
title={{COMI}: Coarse-to-fine Context Compression via Marginal Information Gain},
author={Jiwei Tang and Shilei Liu and Zhicheng Zhang and Yujin Yuan and Libin Zheng and Wenbo Su and Bo Zheng},
booktitle={The Fourteenth International Conference on Learning Representations},
year={2026},
url={https://openreview.net/forum?id=OGDIXDfaN4}
}

@inproceedings{dong2024-mc,
    title = "{MC}-indexing: Effective Long Document Retrieval via Multi-view Content-aware Indexing",
    author = "Dong, Kuicai  and
      Goh Xin Deik, Derrick  and
      Lee, Yi Quan  and
      Zhang, Hao  and
      Li, Xiangyang  and
      Zhang, Cong  and
      Liu, Yong",
    editor = "Al-Onaizan, Yaser  and
      Bansal, Mohit  and
      Chen, Yun-Nung",
    booktitle = "Findings of the Association for Computational Linguistics: EMNLP 2024",
    month = nov,
    year = "2024",
    address = "Miami, Florida, USA",
    publisher = "Association for Computational Linguistics",
    url = "https://aclanthology.org/2024.findings-emnlp.150/",
    doi = "10.18653/v1/2024.findings-emnlp.150",
    pages = "2673--2691",
}

@misc{tang2026readhumancompressingcontext,
      title={Read As Human: Compressing Context via Parallelizable Close Reading and Skimming}, 
      author={Jiwei Tang and Shilei Liu and Zhicheng Zhang and Qingsong Lv and Runsong Zhao and Tingwei Lu and Langming Liu and Haibin Chen and Yujin Yuan and Hai-Tao Zheng and Wenbo Su and Bo Zheng},
      year={2026},
      eprint={2602.01840},
      archivePrefix={arXiv},
      primaryClass={cs.CL},
      url={https://arxiv.org/abs/2602.01840}, 
}

@inproceedings{li-etal-2025-coir,
    title = "{C}o{IR}: A Comprehensive Benchmark for Code Information Retrieval Models",
    author = "Li, Xiangyang  and
      Dong, Kuicai  and
      Lee, Yi Quan  and
      Xia, Wei  and
      Zhang, Hao  and
      Dai, Xinyi  and
      Wang, Yasheng  and
      Tang, Ruiming",
    editor = "Che, Wanxiang  and
      Nabende, Joyce  and
      Shutova, Ekaterina  and
      Pilehvar, Mohammad Taher",
    booktitle = "Proceedings of the 63rd Annual Meeting of the Association for Computational Linguistics (Volume 1: Long Papers)",
    month = jul,
    year = "2025",
    address = "Vienna, Austria",
    publisher = "Association for Computational Linguistics",
    url = "https://aclanthology.org/2025.acl-long.1072/",
    doi = "10.18653/v1/2025.acl-long.1072",
    pages = "22074--22091",
    ISBN = "979-8-89176-251-0",
}

@inproceedings{
gan2025towards,
title={Towards a Theoretical Understanding of Synthetic Data in {LLM} Post-Training: A Reverse-Bottleneck Perspective},
author={Zeyu Gan and Yong Liu},
booktitle={The Thirteenth International Conference on Learning Representations},
year={2025},
url={https://openreview.net/forum?id=UxkznlcnHf}
}

@misc{dong2025docresearcher,
      title={Doc-Researcher: A Unified System for Multimodal Document Parsing and Deep Research}, 
      author={Kuicai Dong and Shurui Huang and Fangda Ye and Wei Han and Zhi Zhang and Dexun Li and Wenjun Li and Qu Yang and Gang Wang and Yichao Wang and Chen Zhang and Yong Liu},
      year={2025},
      eprint={2510.21603},
      archivePrefix={arXiv},
      primaryClass={cs.IR},
      url={https://arxiv.org/abs/2510.21603}, 
}

@article{shumailov2024ai,
  title={AI models collapse when trained on recursively generated data},
  author={Shumailov, Ilia and Shumaylov, Zakhar and Zhao, Yiren and Papernot, Nicolas and Anderson, Ross and Gal, Yarin},
  journal={Nature},
  volume={631},
  number={8022},
  pages={755--759},
  year={2024},
  publisher={Nature Publishing Group UK London}
}

@misc{zhang2026lengthadaptivenetworkbalancinglong,
      title={Length-Adaptive Interest Network for Balancing Long and Short Sequence Modeling in CTR Prediction}, 
      author={Zhicheng Zhang and Zhaocheng Du and Jieming Zhu and Jiwei Tang and Fengyuan Lu and Wang Jiaheng and Song-Li Wu and Qianhui Zhu and Jingyu Li and Hai-Tao Zheng and Zhenhua Dong},
      year={2026},
      eprint={2601.19142},
      archivePrefix={arXiv},
      primaryClass={cs.AI},
      url={https://arxiv.org/abs/2601.19142}, 
}

@inproceedings{ren2025fewshotllmsyntheticdata,
author = {Ren, Jiyuan and Du, Zhaocheng and Wen, Zhihao and Jia, Qinglin and Dai, Sunhao and Wu, Chuhan and Dong, Zhenhua},
title = {Few-shot LLM Synthetic Data with Distribution Matching},
year = {2025},
isbn = {9798400713316},
publisher = {Association for Computing Machinery},
address = {New York, NY, USA},
url = {https://doi.org/10.1145/3701716.3715245},
doi = {10.1145/3701716.3715245},
booktitle = {Companion Proceedings of the ACM on Web Conference 2025},
pages = {432–441},
numpages = {10},
location = {Sydney NSW, Australia},
series = {WWW '25}
}

@misc{tang2026positionbiasshiftingcontext,
      title={Beyond Position Bias: Shifting Context Compression from Position-Driven to Semantic-Driven}, 
      author={Jiwei Tang and Zhijing Huang and Xinyu Zhang and Chen Jason Zhang and Jianxing Yu and Libin Zheng and Rui Meng and Jian Yin},
      year={2026},
      eprint={2605.09463},
      archivePrefix={arXiv},
      primaryClass={cs.CL},
      url={https://arxiv.org/abs/2605.09463}, 
}

\appendix
\section{Detailed LLM Diagnostic Results}
\label{app:llm_diagnosis}

This section provides comprehensive details of the LLM-based diagnostic framework and its findings.

\subsection{Diagnostic Protocol Details}

\paragraph{Sample Collection.}
We collected 1{,}965 matched pairs of ``good'' and ``bad'' generated negatives. Good samples are defined as generated negatives that cluster with queries, positives, or mined negatives in embedding space. Bad samples are those forming pure (source-exclusive) clusters containing only generated negatives.

\paragraph{Stage 1: Hypothesis Discovery.}
We prompted an LLM across three iterative rounds to identify distributional differences between good and bad samples, yielding 49 initial hypotheses categorized into five dimensions: style, domain, noise, structure, and query relation.

\paragraph{Stage 2: Hypothesis Validation.}
Each hypothesis was tested on an independent validation set of 500 sample pairs. We retained hypotheses meeting both criteria: (1) a prevalence difference of $\Delta \geq 10\%$ between bad and good samples, and (2) at least 20\% coverage (the characteristic appears in at least 20\% of samples).

\paragraph{Stage 3: Fine-grained Attribute Annotation.}
For the validated hypotheses, we designed a structured annotation schema covering eight attribute dimensions, enabling quantitative comparison of attribute distributions across 600 labeled samples.

\paragraph{Stage 4: Diversity and Template Analysis.}
We analyzed the diversity characteristics of good and bad samples, finding that the diversity of generated negatives is notably lower than that of samples in the original corpus. Bad samples in particular exhibit pronounced template patterns.

\subsection{Detailed Failure Mode Analysis}

\paragraph{F1: Content Over-generalization ($\Delta=25.0\%$).}
Bad samples exhibit a pronounced tendency toward generic, encyclopedic content (88.8\% in bad vs.\ 63.8\% in good samples). Specific manifestations include background introductions without query-specific information, abstract definitions rather than concrete instances, and absence of specific numbers, dates, or named entities.

\paragraph{F2: Domain Mismatch and Topic Drift ($\Delta=14.0\%$).}
Bad samples exhibit higher rates of domain mismatch, where generated content poorly aligns with the core entities, domain, or situational context of the query.

\paragraph{F3: Semantic Isolation and Noise ($\Delta=12.8\%$).}
Bad samples more frequently contain irrelevant noise, redundant content, or fail to directly address the query, leading to semantic isolation in embedding space.

\paragraph{F4: Core Information Avoidance ($\Delta=11.2\%$).}
Bad samples tend to circumvent the core definitional or direct-answer requirements of the query, instead paraphrasing related background, phenomena, or controversies.

\paragraph{F5: Positive Anchoring Effects.}
Good samples contain significantly more ``wrong entity / wrong time'' type negatives ($+15.0\%$), while bad samples over-represent ``background only'' negatives ($+11.0\%$). This indicates that bad samples tend to generate content highly similar to the positive but merely negating keywords, rather than constructing semantically proximate alternatives with distinct entities or temporal references.

\paragraph{F6: Stylistic Distribution Anomalies.}
Bad samples cluster at ``medium'' specificity levels ($+13.3\%$), while good samples achieve ``very specific'' ratings more frequently ($+14.0\%$). Template analysis reveals pronounced structural patterns in bad samples (e.g., ``topic introduction $+$ background description $+$ negative qualifier''), producing text that is anomalously uniform, overly formal, and structurally regular compared with real corpus data.

\section{Embedding Space Comparison: Vanilla vs.\ CoT Generation}
\label{app:embedding_comparison}

To complement the pure cluster analysis in Section~\ref{sec:validation}, we provide a comprehensive visual comparison of the embedding space geometry under vanilla and CoT-guided generation.

\begin{figure}[t]
    \centering
    \begin{subfigure}{0.48\columnwidth}
        \centering
        \includegraphics[width=\linewidth]{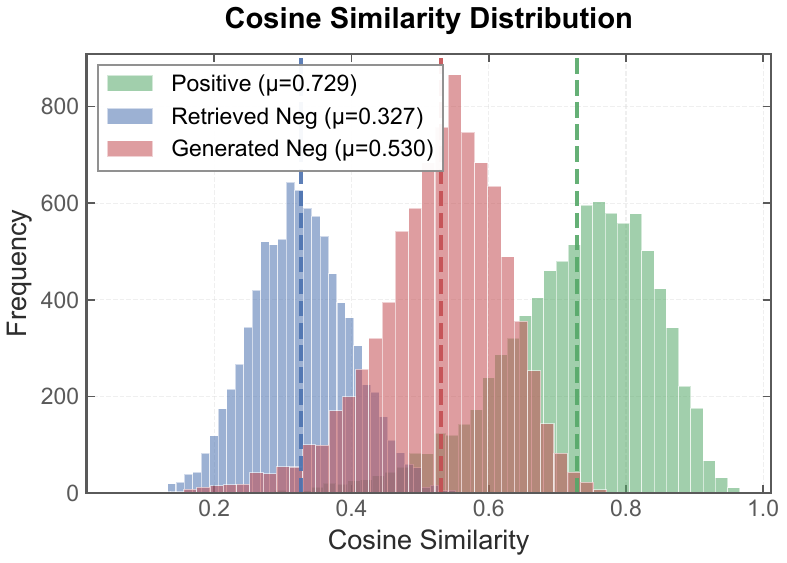}
        \caption{Similarity distribution}
        \label{fig:cot_similarity}
    \end{subfigure}
    \hfill
    \begin{subfigure}{0.48\columnwidth}
        \centering
        \includegraphics[width=\linewidth]{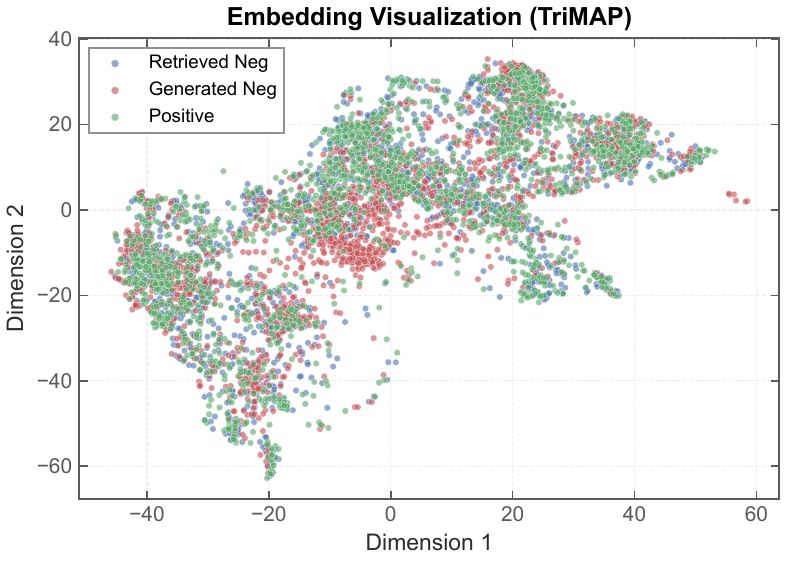}
        \caption{TriMAP visualization}
        \label{fig:cot_clustering}
    \end{subfigure}
    \caption{Embedding space analysis of \method generated negatives: (a) cosine similarity distribution; (b) TriMAP visualization with HDBSCAN cluster assignments. Compared with vanilla generation (Figures~\ref{fig:naive_similarity} and~\ref{fig:naive_clustering}), CoT-generated negatives integrate more naturally with other document types.}
    \label{fig:cot_embedding}
\end{figure}

\paragraph{Similarity distribution.}
Figure~\ref{fig:cot_embedding}(a) shows the cosine similarity distribution between queries and CoT-generated negatives. Compared with the vanilla distribution (Figure~\ref{fig:naive_similarity} in the main text), the CoT distribution more closely resembles that of mined negatives: the spread is wider, the mode is less concentrated, and the tail behavior aligns with real corpus data. This indicates that CoT-guided generation avoids the distributional collapse observed in vanilla generation, where a large fraction of negatives cluster at a narrow similarity range.

\paragraph{Cluster structure.}
Figure~\ref{fig:cot_embedding}(b) presents the TriMAP visualization with HDBSCAN cluster assignments for CoT-generated negatives. In contrast to the vanilla case (Figure~\ref{fig:naive_clustering}), where approximately 24\% of generated negatives form pure source-exclusive clusters, the CoT-generated negatives distribute across clusters together with queries, positives, and mined negatives. No prominent pure clusters of generated negatives are visible, confirming the quantitative finding reported in Section~\ref{sec:validation} that the pure cluster ratio drops below 5\%.

\begin{figure*}[t]
    \centering
    \includegraphics[width=\textwidth]{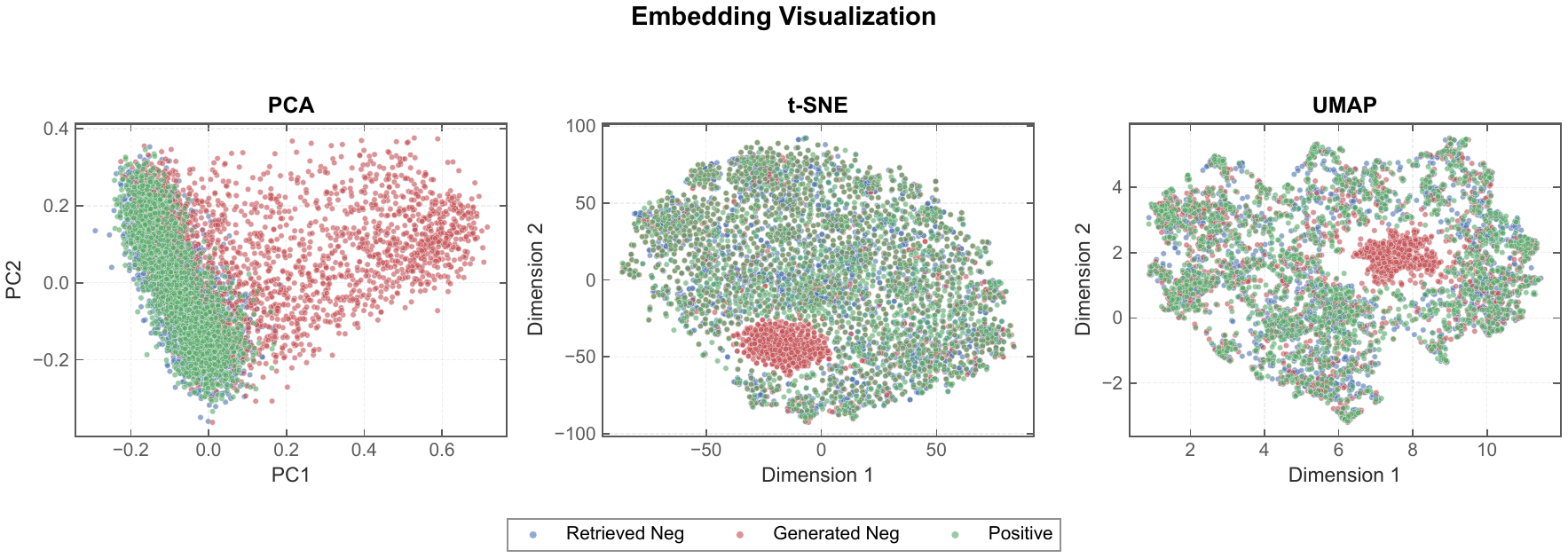}
    \caption{PCA, t-SNE, and UMAP visualization of document embeddings for vanilla generated negatives. Generated negatives (red) form pure clusters absent from other document types across all three projection methods.}
    \label{fig:vanilla_visualization}
\end{figure*}

\begin{figure*}[t]
    \centering
    \includegraphics[width=\textwidth]{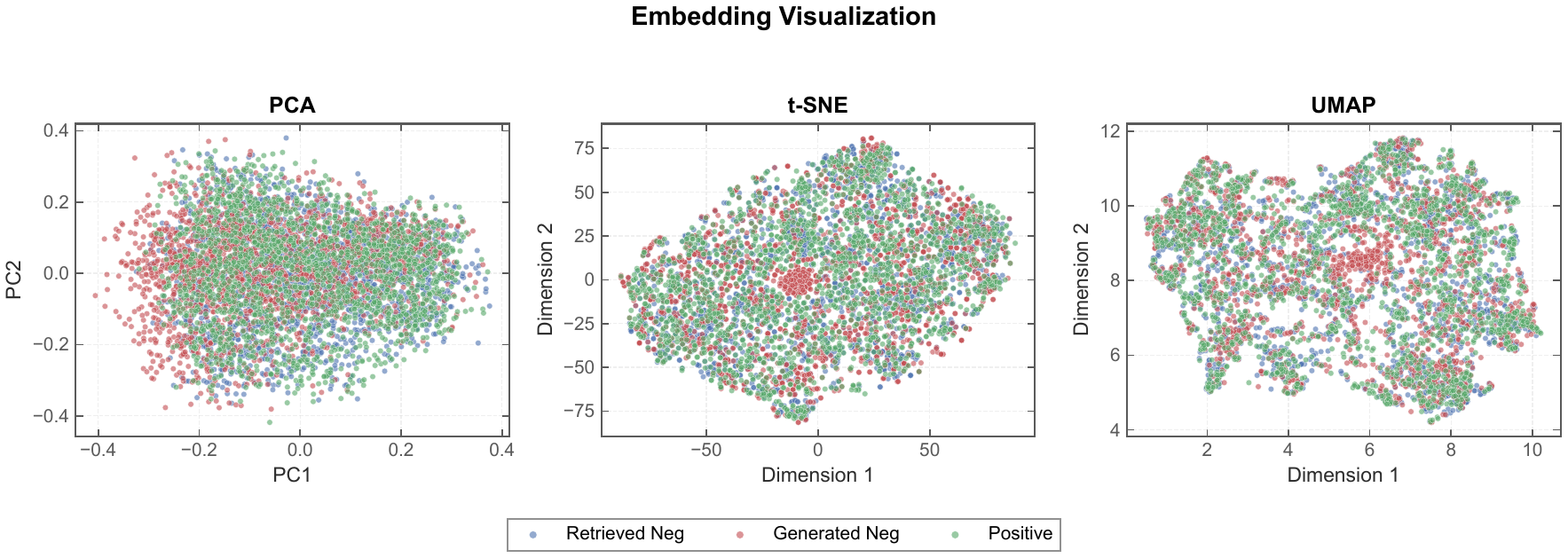}
    \caption{PCA, t-SNE, and UMAP visualization for \method generated negatives. The pure cluster phenomenon is largely eliminated; generated negatives intermix with other document types across all projections.}
    \label{fig:cot_visualization}
\end{figure*}

\paragraph{Multi-method visualization.}
Figures~\ref{fig:vanilla_visualization} and~\ref{fig:cot_visualization} provide a broader view using three complementary dimensionality reduction methods (PCA, t-SNE, and UMAP). Under vanilla generation (Figure~\ref{fig:vanilla_visualization}), all three methods consistently reveal isolated regions populated exclusively by generated negatives, confirming that the pure cluster phenomenon is not an artifact of a particular visualization technique. Under CoT-guided generation (Figure~\ref{fig:cot_visualization}), these isolated regions largely disappear: generated negatives intermix with other document types across all three projections. This convergence across independent visualization methods strengthens the conclusion that CoT-guided generation produces negatives whose distributional properties are consistent with real corpus data, thereby removing the source-specific fingerprint that enables shortcut learning.

\section{Generation Prompts}
\label{app:prompts}

We use a strong proprietary LLM as the generation model. The pipeline consists of two LLM calls per query: Step 2 (information requirement decomposition and disruption strategy design) and Step 3 (constrained hard negative generation). Step 1 (relevance definition and writing style) is prepared offline for each dataset and injected into both prompts. We present condensed versions of the prompt templates below; dataset-specific relevance definitions and writing style constraints are omitted for brevity. Placeholders are shown in \texttt{\{braces\}}.

\paragraph{Step 2: Information Requirement Decomposition.}
This prompt instructs the LLM to analyze the relationship between a query and its positive document, decompose the information need into a structured reasoning chain, and design targeted disruption strategies for each critical node.

\begin{figure}[h]
\centering
\fbox{\parbox{0.93\columnwidth}{\footnotesize
\texttt{You are an information retrieval expert responsible for analyzing the query's information need structure and designing ``information need disruption'' strategies to generate hard negatives.}\\[2pt]
\texttt{\{Dataset-specific relevance definition\}}\\[2pt]
\texttt{Query: \{query\}}\\
\texttt{Positive Document: \{pos\_doc\}}\\[2pt]
\texttt{Task:}\\
\texttt{Part 1: Analyze the Relationship}\\
\texttt{1. Query information need (one sentence)}\\
\texttt{2. How positive sample satisfies it (one sentence)}\\
\texttt{3. Answer boundary: what counts as answering}\\
\texttt{4. Chain of Thought: Decompose into 4--8 reasoning nodes, each with facet\_type (information\_need / entity / attribute / constraint / reasoning / style), content, and critical flag.}\\[2pt]
\texttt{Part 2: Design Disruption Strategies}\\
\texttt{For each critical node, design 2--3 strategies:}\\
\texttt{- entity\_shift: entity replacement}\\
\texttt{- intent\_drift: information need drift}\\
\texttt{- constraint\_violation: breaking constraints}\\
\texttt{- upstream\_downstream: scope shift}\\[2pt]
\texttt{Output: JSON with query\_info and chain\_nodes.}
}}
\caption{Condensed Step 2 prompt template for information requirement decomposition and disruption strategy design.}
\label{fig:prompt_step2}
\end{figure}

\paragraph{Step 3: Constrained Hard Negative Generation.}
This prompt takes the structured output of Step 2 and generates one hard negative document per disruption strategy, constrained to match the corpus writing style.

\begin{figure}[h]
\centering
\fbox{\parbox{0.93\columnwidth}{\footnotesize
\texttt{You are an information retrieval expert responsible for constructing hard negative samples for dense retrieval models.}\\[2pt]
\texttt{\{Dataset-specific relevance definition\}}\\[2pt]
\texttt{Query: \{query\}}\\
\texttt{Positive Document: \{pos\_doc\}}\\
\texttt{Query Analysis: \{query\_info from Step 2\}}\\
\texttt{Chain Nodes and Strategies: \{chain\_nodes from Step 2\}}\\
\texttt{Candidate Negatives (from corpus, for style reference): \{top 3 mined negatives\}}\\[2pt]
\texttt{Task: For each disruption strategy, generate 1 hard negative.}\\[2pt]
\texttt{Core Requirements:}\\
\texttt{1. Cannot answer the query in any form (correct, incorrect, or partial).}\\
\texttt{2. Prioritize rewriting candidate negatives; preserve their style and noise.}\\
\texttt{3. Only generate from scratch if no candidates fit; simulate real corpus style.}\\[2pt]
\texttt{\{Dataset-specific writing style constraints\}}\\[2pt]
\texttt{Output: JSON with node\_id, strategy\_id, generated text, and break\_explanation.}
}}
\caption{Condensed Step 3 prompt template for constrained hard negative generation.}
\label{fig:prompt_step3}
\end{figure}

\paragraph{Dataset-specific configurations.}
For each dataset, we provide tailored relevance definitions and writing style requirements. The relevance definition specifies what constitutes a positive versus a negative document in the context of that dataset (e.g., for HotpotQA, positives must provide bridging information for multi-hop reasoning; for TriviaQA, positives must address the trivia entity being asked about). The writing style section constrains the generation to match the stylistic properties of the real corpus (e.g., Wikipedia encyclopedic style for NQ, mixed web and Wikipedia style for TriviaQA, entity-centric passages with attribute density for HotpotQA).

\section{Hyperparameter Sensitivity and Selection Strategy}
\label{app:sensitivity}

\begin{figure}[t]
    \centering
    \includegraphics[width=\columnwidth]{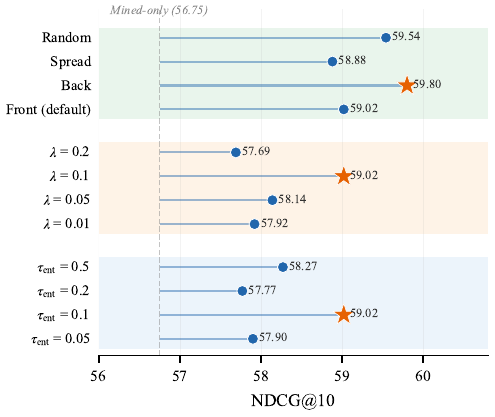}
    \caption{Hyperparameter and selection strategy robustness on TQA. All configurations of $\tau_{\text{ent}}$, $\lambda$, and chain position selection outperform the Mined-only baseline (56.75, vertical dashed line). Stars indicate the optimal configuration within each group.}
    \label{fig:sensitivity_appendix}
\end{figure}

We conduct a systematic sensitivity analysis on TQA across the entropy regularization hyperparameters ($\tau_{\text{ent}}$, $\lambda$) and the chain position selection strategy. The Mined-only baseline achieves 56.75 NDCG@10 on this dataset. Figure~\ref{fig:sensitivity_appendix} summarizes all configurations; every tested setting substantially outperforms the baseline.

\paragraph{Entropy temperature $\tau_{\text{ent}}$.}
Varying $\tau_{\text{ent}} \in \{0.05, 0.1, 0.2, 0.5\}$ while fixing $\lambda = 0.1$, the optimal value is $\tau_{\text{ent}} = 0.1$ (59.02 NDCG@10). Both extremes degrade performance: an overly sharp distribution ($\tau_{\text{ent}} = 0.05$, 57.90) amplifies noise in similarity estimates, while an overly flat one ($\tau_{\text{ent}} = 0.5$, 58.27) weakens the regularization signal. Nonetheless, all tested values improve over the baseline, indicating that the method is not sensitive to precise temperature tuning.

\paragraph{Entropy loss weight $\lambda$.}
Varying $\lambda \in \{0.01, 0.05, 0.1, 0.2\}$ while fixing $\tau_{\text{ent}} = 0.1$, the optimal value is $\lambda = 0.1$ (59.02). Both under-regularization ($\lambda = 0.01$, 57.92; insufficient suppression of shortcuts) and over-regularization ($\lambda = 0.20$, 57.69; interference with the contrastive objective) lead to suboptimal results. All values improve over the baseline, confirming the robustness of the approach.

\paragraph{Chain position selection.}
When selecting $k = 3$ negatives from different positions in the reasoning chain, the ``Back'' strategy (59.80 NDCG@10) outperforms the default ``Front'' selection (59.02), suggesting that late-chain perturbations---which target fine-grained constraints while preserving topical proximity---produce harder and more effective training negatives. ``Random'' selection (59.54) also outperforms the default, indicating that diversity in perturbation types is beneficial. Overall, all selection strategies substantially outperform both Mined-only (56.75) and SyNeg (53.16), confirming the CoT generation framework's robustness to the specific selection policy.

\section{Detailed Experimental Setup}
\label{app:experiments}

\begin{table}[h]
    \centering
    \caption{Complete training hyperparameters for all experiments.}
    \label{tab:hyperparams}
    \begin{tabular}{ll}
        \toprule
        \multicolumn{2}{l}{\textit{Infrastructure}} \\
        \midrule
        Base model & Qwen3-0.6B \\
        Training framework & Swift \\
        Training data source & BGE-M3-Data \\
        Hardware & 8 $\times$ consumer-grade 24GB accelerator \\
        Generation model & Proprietary LLM API \\
        \midrule
        \multicolumn{2}{l}{\textit{Optimization}} \\
        \midrule
        Learning rate & $5 \times 10^{-5}$ \\
        Per-device batch size & 4 \\
        Gradient accumulation steps & 8 \\
        Effective batch size & 256 \\
        Max sequence length & 8{,}192 \\
        Training epochs & 10 (early stopping, patience 3) \\
        Warmup ratio & 0.1 \\
        \midrule
        \multicolumn{2}{l}{\textit{Contrastive Learning \& Entropy Regularization}} \\
        \midrule
        InfoNCE temperature $\tau$ & 0.05 \\
        Entropy temperature $\tau_{\text{ent}}$ & 0.1 \\
        Entropy loss weight $\lambda$ & 0.05 \\
        Mined negatives per query & 15 \\
        Generated negatives per query & 3 (default; up to 10 in analysis) \\
        \bottomrule
    \end{tabular}
\end{table}

Table~\ref{tab:hyperparams} lists the complete training configuration used across all experiments. We fine-tune Qwen3-0.6B~\cite{qwen3embedding} with the Swift framework~\cite{swift} on training data derived from BGE-M3-Data~\cite{chen-etal-2024-m3}. Hard negatives are mined via BM25 retrieval from the original corpus, retaining 15 negatives per query. For \method, CoT-guided negatives are generated using a strong proprietary LLM API through a two-call pipeline (decomposition + generation); 3 negatives per query are used by default for fair comparison with vanilla generation~\cite{li2024synegllmdrivensynthetichardnegatives}. All experiments are conducted on 8 consumer-grade 24GB accelerators with full-parameter training and early stopping (patience 3 epochs). The entropy regularization hyperparameters ($\tau_{\text{ent}} = 0.1$, $\lambda = 0.1$) are selected based on the sensitivity analysis in Appendix~\ref{app:sensitivity}.

\subsection{Dataset Details}
\label{app:datasets}

\paragraph{Data sources.}
All training data are derived from BGE-M3-Data~\cite{chen-etal-2024-m3}, an open-source multilingual retrieval dataset collection. For MS~MARCO, we use mMARCO-zh~\cite{Bonifacio2021MMarco}, a Chinese translation of the original English MS~MARCO dataset~\cite{marco}. HotpotQA, NQ, and TriviaQA retain their original English versions.

\paragraph{Training data statistics.}
Table~\ref{tab:train_stats} reports the number of training queries per dataset. For each query, we retrieve 15 hard negatives via BM25 from the original corpus (after excluding positive documents). Random negatives are sampled uniformly from the same corpus pool (up to 100K documents per dataset) with positives removed.

\begin{table}[h]
    \centering
    \caption{Training data statistics (number of queries).}
    \label{tab:train_stats}
    \begin{tabular}{lc}
        \toprule
        Dataset & Training Queries \\
        \midrule
        mMARCO-zh & 8,734 \\
        NQ & 9,576 \\
        TriviaQA & 9,579 \\
        HotpotQA & 9,578 \\
        \bottomrule
    \end{tabular}
\end{table}

\paragraph{Evaluation protocol.}
We adopt the evaluation framework from Qwen3-Embedding~\cite{qwen3embedding}. For datasets with large-scale evaluation sets, we sample 3K queries and 100K corpus documents to ensure computational tractability. Table~\ref{tab:eval_stats} summarizes the evaluation statistics.

\begin{table}[h]
    \centering
    \caption{Evaluation data statistics.}
    \label{tab:eval_stats}
    \begin{tabular}{lccccc}
        \toprule
        Dataset & Queries & Corpus & Qrels & Split & Lang \\
        \midrule
        mMARCO-zh & 6,980 & 106,813 & 7,437 & dev & zh \\
        HotpotQA & 1,000 & 225,621 & 2,000 & test & en \\
        TriviaQA & 3,000 & 100,000 & 5,239 & test & en \\
        NQ & 3,000 & 100,000 & 3,644 & test & en \\
        \bottomrule
    \end{tabular}
\end{table}

\paragraph{Baseline configurations.}
Table~\ref{tab:baseline_config} details the negative sampling configuration for each baseline method. Vanilla generation refers to the standard LLM-based hard negative generation approach (i.e., SyNeg~\cite{li2024synegllmdrivensynthetichardnegatives}) without structured reasoning or explicit relevance modeling.

\begin{table}[h]
    \centering
    \caption{Baseline negative sampling configurations.}
    \label{tab:baseline_config}
    \begin{tabular}{lcc}
        \toprule
        Method & Mined Neg. & Generated Neg. \\
        \midrule
        Random & 0 (15 Random) & 0 \\
        Mined-only & 15 (BM25) & 0 \\
        Vanilla Generation & 0 & up to 15 (Vanilla) \\
        CoT Generation & 0 & up to 15 (CoT) \\
        SyNeg & 15 (BM25) & 3 (Vanilla) \\
        \method & 15 (BM25) & 3 (CoT) \\
        \bottomrule
    \end{tabular}
\end{table}

\section{Limitations}
\label{sec:limitations}

Our work has several limitations. First, due to compute resource constraints, all experiments are conducted with a single base model architecture (Qwen3-0.6B); validating whether the identified phenomena and proposed solutions generalize to larger encoder models remains important future work. Second, API resource constraints limit the scale of data generation and the scope of fine-tuning experiments. Third, the CoT-guided generation pipeline relies on a capable LLM and incurs Inference or API costs that may limit scalability in cost-sensitive settings, particularly when processing lengthy documents. Recent efforts have highlighted the broader challenges of length-induced biases in long context modeling~\cite{dong2024-mc}, such as attention polarization in sequential recommendation~\cite{zhang2026lengthadaptivenetworkbalancinglong}, code retrieval~\cite{li-etal-2025-coir}, and multimodal retrieval~\cite{dong2025mmdocir,dong2025mmdocrag}. Long context compression techniques~\cite{tang-etal-2025-perception,tang2026gmsaenhancingcontextcompression,tang2026comi, tang2026readhumancompressingcontext,tang2026positionbiasshiftingcontext} present a promising avenue for document preprocessing~\cite{dong2025docresearcher}, minimizing token overhead while safeguarding essential semantic information. Future research could leverage these methods to manage complex, lengthy context during negative synthesis, thereby optimizing end-to-end latency without compromising performance. Finally, the distributional artifacts identified in this study are specific to the generation model and prompting strategy used; different LLMs or generation paradigms may introduce different types of artifacts, and the extent to which our entropy regularization generalizes to these settings warrants further investigation.

\end{document}